\documentclass[runningheads]{llncs}

\usepackage{eccv}

\usepackage{eccvabbrv}

\usepackage{graphicx}
\usepackage{booktabs}

\usepackage[accsupp]{axessibility}  % Improves PDF readability for those with disabilities.

\usepackage[dvipsnames]{xcolor}
\usepackage{multirow}
\usepackage{multicol}
\usepackage{makecell}
\usepackage{algorithm}
\usepackage{algpseudocode}
\usepackage{subcaption}
\usepackage{balance}
\usepackage{colortbl}
\usepackage{adjustbox} 
\usepackage{setspace}
\usepackage{wrapfig}

\definecolor{forestgreen}{RGB}{34,139,34}
\definecolor{colorTab}{rgb}{0.92,0.95,0.92}

\newcommand\blfootnote[1]{%
  \begingroup
  \renewcommand\thefootnote{}\footnote{#1}%
  \addtocounter{footnote}{-1}%
  \endgroup
}

\usepackage[pagebackref,breaklinks,colorlinks,citecolor=eccvblue]{hyperref}
\usepackage{orcidlink}

\begin{document}

\title{DPoser: Diffusion Model as \\ Robust 3D Human Pose Prior} 



\author{%
	Junzhe Lu$^{1}$, Jing Lin$^{2}$, Hongkun Dou$^{1}$, \\ Ailing Zeng$^{3}$, Yue Deng$^{1}$, Yulun Zhang$^{4}$, and Haoqian Wang$^{2}$ \\
 $^1$ Beihang University, $^{2}$ Tsinghua University, \\ $^{3}$ International Digital Economy Academy (IDEA), $^{4}$ ETH Zürich \\
    \url{https://dposer.github.io}
}

\authorrunning{J.~Lu et al.}

\institute{}

\maketitle

\begin{abstract}
  This work targets to construct a robust human pose prior. However, it remains a persistent challenge due to biomechanical constraints and diverse human movements. Traditional priors like VAEs and NDFs often exhibit shortcomings in realism and generalization, notably with unseen noisy poses. To address these issues, we introduce DPoser, a robust and versatile human pose prior built upon diffusion models. 
  DPoser regards various pose-centric tasks as inverse problems and employs variational diffusion sampling for efficient solving. Accordingly, designed with optimization frameworks, DPoser seamlessly benefits human mesh recovery, pose generation, pose completion, and motion denoising tasks. Furthermore, due to the disparity between the articulated poses and structured images, we propose truncated timestep scheduling to enhance the effectiveness of DPoser. 
  Our approach demonstrates considerable enhancements over common uniform scheduling used in image domains, boasting improvements of 5.4\%, 17.2\%, and 3.8\% across human mesh recovery, pose completion, and motion denoising, respectively.
  Comprehensive experiments demonstrate the superiority of DPoser over existing state-of-the-art pose priors across multiple tasks. 
\vspace{0pt}
\blfootnote{Corresponding authors: Yulun Zhang and Haoqian Wang.}
\end{abstract}

\section{Introduction}
\label{sec:intro}

Accurate modeling of human pose is a fundamental research topic that can benefit various applications, from human-robot interaction to augmented and virtual reality experiences. 
Many real-world applications rely on a prior distribution of valid human poses to perform tasks like body model fitting, motion capture, and gesture recognition. The complexity of human biomechanics, coupled with the extensive kinematic variability in movement patterns, presents a significant challenge in constructing a robust and realistic human pose prior.

Previous efforts to model human pose prior have mainly employed techniques such as Gaussian Mixture Models (GMMs)~\cite{bogo2016keep}, Variational Autoencoders (VAEs)~\cite{pavlakos2019expressive}, and Neural Distance Fields (NDFs)~\cite{tiwari2022pose}.
Each technique, however, faces its own set of limitations. GMMs, for instance, might lead to the generation of implausible poses due to their unbounded nature. 
VAEs, restricted by their Gaussian assumptions, tend to generate average poses that may not accurately capture the full spectrum of human actions.
Meanwhile, NDFs have shown promise in 3D surface modeling but struggle with generalizing across the complex, high-dimensional landscape of human pose manifolds.
These limitations highlight a pressing need for a more comprehensive and dependable approach to modeling human pose priors, an endeavor this work seeks to address.

Recently, Diffusion models~\cite{ho2020denoising, song2020score, dhariwal2021diffusion, karras2022elucidating} have gained traction for their prowess in capturing complex, high-dimensional data distributions and enabling versatile sampling techniques. 
Their application has been seen in generating lifelike human motion sequences~\cite{zhao2023modiff, shafir2023human} and functioning as multi-hypothesis pose estimators from 2D inputs~\cite{holmquist2023diffpose, ci2023gfpose}.
However, these models are designed for specific generation tasks or tailored to work with conditional input data, which limits their applicability in broader contexts. 
% such as inverse kinematics and motion denoising
The potential of diffusion models as a universal human pose prior remains largely untapped, and effective optimization methods for diverse tasks remain unanswered.

\begin{figure}[!t]
\centering
\includegraphics[width=\textwidth]{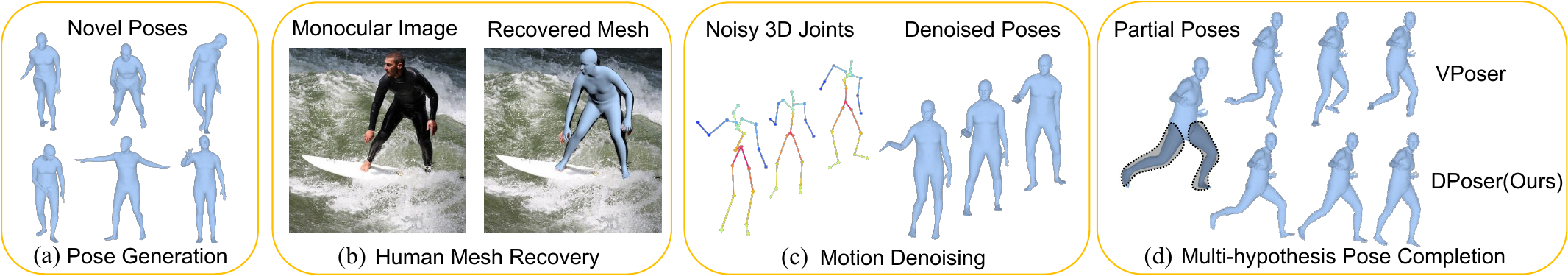}
\caption{An overview of DPoser's versatility and performance across multiple pose-related tasks. Built on diffusion models, DPoser serves as a robust and adaptable pose prior. Shown are scenarios in (a) pose generation, (b) human mesh recovery, (c) motion denoising, and (d) pose completion. DPoser consistently outstrips existing priors like VPoser~\cite{pavlakos2019expressive} in performance benchmarks.}
\label{fig:teaser}
\vspace{-5mm}
\end{figure}

In this work, we propose DPoser, a novel approach that leverages time-dependent denoiser learned from expansive motion capture datasets to construct a robust human pose prior. 
We regard various pose-centric tasks as inverse problems and suggest the integration of DPoser via variational diffusion sampling techniques~\cite{mardani2023variational} as a regularization component within optimization frameworks like SMPLify~\cite{bogo2016keep}.
Furthermore, our investigations reveal that significant pose-related information during diffusion is predominantly located at the latter stages of the diffusion trajectory. This revelation inspired us to develop a novel truncated timestep scheduling strategy for optimization. Our method outperforms the standard uniform scheduling, showing gains of 5.4\%, 17.2\%, and 3.8\% in human mesh recovery, pose completion, and motion denoising, respectively.

In summary, our main contributions are as follows:
\begin{itemize}
\vspace{0.5mm}
\item We introduce DPoser, a novel framework based on diffusion models to craft a robust and flexible human pose prior, geared for seamless integration across diverse pose-related tasks via test-time optimization.
\vspace{0.5mm}
\item We analyze the impact of diffusion timesteps in the pose domain and propose truncated scheduling for more efficient optimization.

\vspace{0.5mm}
\item Through extensive experiments, we establish that DPoser outshines state-of-the-art (SOTA) pose priors in a variety of downstream tasks.

\end{itemize}

\section{Related Work}

\subsection{Human Pose Priors}
Human body models such as SMPL~\cite{loper2015smpl} serve as powerful tools for parameterizing both pose and shape, thereby offering a comprehensive framework for describing human gestures. Within the SMPL model, body poses are captured using rotation matrices or joint angles linked to a kinematic skeleton. Adjusting these parameters enables the representation of a diverse range of human actions. Nonetheless, feeding unrealistic poses into these models can result in non-viable human figures, primarily because plausible human poses are confined within a complex, high-dimensional manifold due to biomechanical constraints.

Various strategies~\cite{bogo2016keep, pavlakos2019expressive, tiwari2022pose, davydov2022adversarial} have been put forward to build human pose priors. Generative frameworks like GMMs, VAEs~\cite{kingma2013auto}, and Generative Adversarial Networks 
 (GANs)~\cite{creswell2018generative} have shown promise in encapsulating the multifaceted pose distribution, facilitating advancements in tasks like human mesh recovery~\cite{kanazawa2018end, georgakis2020hierarchical}. 
Further, some studies have delved into conditional pose priors tailored to specific tasks, incorporating extra information such as image features~\cite{qiu2023learning, cho2023generative}, 2D joint coordinates~\cite{ci2023gfpose}, or sequences of preceding poses~\cite{ling2020character, rempe2021humor}.
Our initiative leans towards an unconditional pose prior approach, training DPoser on extensive motion capture data without relying on additional inputs like images or text, aiming for a versatile application across various pose-related scenarios.

\subsection{Diffusion Models for Pose-centric Tasks}
\vspace{-1mm}
Diffusion models~\cite{song2019generative, song2020score, ho2020denoising, song2020denoising} have emerged as powerful tools for capturing intricate data distributions, aligning particularly well with the demands of multi-hypothesis estimation in ambiguous human poses.
Notable works include DiffPose~\cite{holmquist2023diffpose}, which leverages a Gaussian Mixture Model-guided forward diffusion process~\cite{nachmani2021non} and employs a Graph Convolutional Network (GCN)~\cite{kipf2016semi} architecture conditioned on 2D pose sequences for 3D pose estimation by learned reverse process (\ie, generation).
In a similar vein, DiffusionPose~\cite{qiu2023learning} and GFPose~\cite{ci2023gfpose} employ the generation-based pipeline but take different approaches in conditioning. 
Further, ZeDO~\cite{jiang2023back} concentrates on 2D-to-3D pose lifting, while Diff-HMR~\cite{cho2023generative} and DiffHand~\cite{li2023diffhand} explore estimating SMPL parameters and hand mesh vertices, respectively.
BUDDI~\cite{muller2023generative} stands out for using diffusion models to capture the joint distribution of interacting individuals and leveraging SDS loss~\cite{poole2022dreamfusion, wang2023score} for optimization during testing phases.

While DPoser shares a similar optimization implementation with BUDDI, it sets itself apart by introducing a wider perspective of inverse problems and equipping an innovative timestep scheduling strategy tailored to the characteristics of human poses. 
Unlike other approaches~\cite{jiang2023back, qiu2023learning, ci2023gfpose, holmquist2023diffpose} that primarily focus on 3D location-based representation, DPoser takes on the more demanding task of modeling SMPL-based rotation pose representation. 
This adds complexity due to the intricacies involved in representing rotations, positioning DPoser as a more versatile solution within the realm of pose-centric tasks.

\section{Methods}

\subsection{Preliminary: Score-based Diffusion Models}
Diffusion models~\cite{sohl2015deep, song2019generative, song2020score, ho2020denoising} operationalize generative processes by inverting a predefined forward diffusion process, typically formulated as a linear stochastic differential equation (SDE). Formally, the data trajectory \(\left\{\mathbf{x}(t)\in\mathbb{R}^n\right\}_{t\in[0,1]}\) follows the forward SDE given by:
\begin{equation}
    \mathrm{d} \mathbf{x}= \mu(t) \mathbf{x} \mathrm{d} t+ g(t) \mathrm{d} \mathbf{w},
    \label{eq:forward SDE}
\end{equation}
where \(\mu(t) \mathbf{x} \in \mathbb{R}^n\) and \(g(t) \in \mathbb{R}\) represent the drift and diffusion coefficients, while \(\mathbf{w}\) is a standard Wiener process.

The affine drift coefficients ensure analytically tractable Gaussian perturbation kernels, denoted by \(p_{0t}(\mathbf{x}_t \mid \mathbf{x})=\mathcal{N}(\mathbf{x}_t;\alpha_{t}\mathbf{x},\sigma_{t}^{2}\mathbf{I})\), where the exact coefficients \(\alpha_{t}, \sigma_{t}\) can be obtained with standard techniques~\cite{sarkka2019applied}. Using appropriately designed \(\alpha_{t}\) and \(\sigma_{t}\), this allows the data distribution \(\mathbf{x}_0 \sim p_{data}\) to morph into a tractable isotropic Gaussian distribution \(\mathbf{x}_1 \sim  \mathcal{N}(\mathbf{0},\mathbf{I})\) via forward diffusion.

To recover data distribution \(p_{data}\) from the Gaussian distribution \(\mathcal{N}(\mathbf{0},\mathbf{I})\), we can simulate the corresponding reverse SDE of Eq.~\eqref{eq:forward SDE}~\cite{song2020score}:
\begin{equation}
\mathrm{d}\mathbf{x} = [ \mu(t) \mathbf{x} - g(t)^2 \nabla_{\mathbf{x}_t}\log p_t\left(\mathbf{x}_t\right) ] \mathrm{d}t + g(t) \mathrm{d}\bar{\mathbf{w}}.
\label{eq:reverse SDE}
\end{equation}
The so-called score function~\cite{liu2016kernelized}, \(\nabla_{\mathbf{x}_t}\log p_t\left(\mathbf{x}_t\right)\), serves as an unknown term in Eq.~\eqref{eq:reverse SDE} and can be approximated by a neural network parameterized as \(\epsilon_\phi(\mathbf{x}_t;t) \approx -\sigma_t \nabla_{\mathbf{x}_t}\log p_t\left(\mathbf{x}_t\right)\)\footnote{This parameterization is obtained from the deep connection between the noise prediction in diffusion models and score function estimation in score-based models. We provide a brief recap in the Appendix.}. To learn the score functions, employing denoising score matching techniques~\cite{vincent2011connection}, we perturb the data points with noise as per:
\begin{equation}
    \mathbf{x}_t = \alpha_{t}\mathbf{x}_0+\sigma_{t}\epsilon, \epsilon \sim \mathcal{N}(\mathbf{0},\mathbf{I})
    \label{eq:forward diffusion}.
\end{equation}
Subsequently, feeding \(\mathbf{x}_t\) and \(t\) as input, we train the time-dependent noise predictor \(\epsilon_\phi(\mathbf{x}_t;t)\) using an L2-loss defined as~\cite{ho2020denoising}:
\begin{equation}
    \mathbb{E}_{\mathbf{x}_0\sim p_{\mathrm{data}},\epsilon\sim\mathcal{N}(\mathbf{0}, \mathbf{I}),t\sim\mathcal{U}[0,1]} \left[w(t)||\epsilon-\epsilon_\phi(\mathbf{x}_t;t)||_2^2\right],
    \label{eq:training objective}
\end{equation}
where \(w(t)\) denotes a positive weighting function.

Upon successful training, the score functions can be estimated and used to solve the reverse SDE (Eq.~\eqref{eq:reverse SDE}). 
Through techniques like Euler-Maruyama discretization, we can generate novel samples by simulating the reverse SDE.

\subsection{Learning Pose Prior with Unconditional Diffusion Models}
\noindent \textbf{SMPL-based pose representation.}\
To build a flexible 3D human pose prior, we propose to utilize the SMPL body model~\cite{loper2015smpl}, which can be viewed as a differentiable function \([J, V] = M(\theta,\beta)\) that maps body joint angles \(\theta \in \mathbb{R}^{3\times21}\) and shape parameters \(\beta \in \mathbb{R}^{10}\) to mesh vertices \(V \in \mathbb{R}^{3\times6890} \) and joint positions \(J \in \mathbb{R}^{3\times22} \). Our target is to model the distribution of joint angles \(p(\theta)\). 

\noindent \textbf{Training of unconditional diffusion models.}\
To this end, we adopt an unconditional diffusion model to learn the pose representation \(\theta\). 
This approach aligns with a task-agnostic strategy, focusing solely on the distribution of 3D poses. 
We employ sub-VP SDEs as outlined in \cite{song2020score}, which have demonstrated efficacy in sampling quality, for constructing our diffusion model. Specifically, our chosen forward SDE (Eq.~\eqref{eq:forward SDE}) is given by:
\begin{equation}
    \mathrm{d}\mathbf{x}=-\frac{1}{2}\xi(t)\mathbf{x}\mathrm{d}t+\sqrt{\xi(t)(1-e^{-2\int_0^t\xi(s)\mathrm{d}s})}\mathrm{d}\mathbf{w},
\end{equation}
where \(\xi(t)\) denotes linear scheduled noise scales.
The coefficients needed in Eq.~\eqref{eq:forward diffusion} can be obtained as \(\alpha_{t}=e^{-\frac{1}{2}\int_{0}^{t}\xi(s)\mathrm{d}s},  \sigma_{t}=1-e^{-\int_{0}^{t}\xi(s)\mathrm{d}s}\). 

During training, we initiate with a clean data point \(\mathbf{x}_0\)—essentially, our pose representation \(\theta\)—and introduce noise to generate samples \(\mathbf{x}_t\) according to the forward process detailed in Eq.~\eqref{eq:forward diffusion}.
Then we apply the objective in Eq.~\eqref{eq:training objective} to train the noise predictor \(\epsilon_\phi(\mathbf{x}_t;t)\) with weights \(w(t)=\sigma_{t}^2\) as suggested in \cite{song2020score}.

\subsection{Optimization Leveraging Diffusion Priors}
\label{sec:DPoser}

The acquired score functions or noise predictors, denoted as \(\epsilon_\phi(\mathbf{x}_t;t)\), permit the direct generation of plausible poses through Eq.~\eqref{eq:reverse SDE}. Yet, the broader integration of diffusion priors into general optimization frameworks remains an open avenue. We address this by reframing pose-related tasks as inverse problems and applying variational diffusion sampling techniques~\cite{mardani2023variational} for efficient resolution.

\noindent \textbf{Inverse problem formulation.}\
Consider an original signal \(\mathbf{x}_0\). Inverse problems can be encapsulated by Eq.~\eqref{eq:inverse problem} as:
\vspace{-2mm}
\begin{equation}
    \mathbf{y}=\mathcal{A}(\mathbf{x}_0)+\mathbf{n},\quad \mathbf{y},\mathbf{n}\in\mathbb{R}^d,~\mathbf{x}_0\in\mathbb{R}^n,
    \label{eq:inverse problem}
\vspace{-2mm}
\end{equation}
where \(\mathcal{A}\) symbolizes the measurement operator and \(\mathbf{n}\) constitutes noise, assumed to be white Gaussian \(\mathcal{N}(\mathbf{0}, \sigma_n^2\mathbf{I})\). In the context targeted in this study, \(\mathbf{x}_0\) always refers to body poses in SMPL~\cite{loper2015smpl}. This formulation allows us to approach various pose-centric tasks by adapting \(\mathcal{A}\) and interpreting \(\mathbf{y}\) accordingly:
\begin{itemize}
    \item \textbf{Pose completion}: Here, \(\mathcal{A}\) serves as a mask matrix to simulate partially observed poses, with \(\mathbf{y}\) being the incomplete pose data.
    \item \textbf{Motion denoising}: In this scenario, \(\mathcal{A}\) applies SMPL's forward kinematics, treating \(\mathbf{y}\) as the observed noisy 3D joints.
    \item \textbf{Human mesh recovery}: \(\mathcal{A}\) integrates SMPL's forward kinematics and camera projection to relate \(\mathbf{y}\) to 2D joint observations in images.
\end{itemize}
The aim is to recover the original signal \(\mathbf{x}_0\), where, within the Bayesian framework, our objective shifts to sampling from the posterior distribution \(p\left(\mathbf{x}_0 \mid \mathbf{y}\right)\).

\noindent \textbf{Solving inverse problems with diffusion models.}\
Various techniques~\cite{graikos2022diffusion, kawar2022denoising, chung2022diffusion, chung2022improving, song2022pseudoinverse, mardani2023variational} have been explored to simulate this posterior sampling process based on unconditional diffusion priors \(p\left(\mathbf{x}_0;\phi\right)\). Among them, the sampling-based scheme is widely explored and applied in tasks like image restoration. These methods incorporate the observation information \(\mathbf{y}\) into the generation process of \(\mathbf{x}_0\) through techniques like gradient guidance~\cite{chung2022diffusion, chung2022improving} and back projection~\cite{song2020score, kawar2022denoising, chung2022improving}. However, such methods rooted in generation are inconvenient for handling diverse pose-related tasks. To navigate these challenges, we adopt variational diffusion sampling~\cite{mardani2023variational} to build general optimization frameworks. Specifically, it employs a variational distribution \( q\left(\mathbf{x}_0 \mid \mathbf{y}\right):= \mathcal{N}(\mu, \sigma^2\mathbf{I}) \) and aims to minimize the Kullback-Leibler (KL) divergence between this variational distribution and the true posterior, mathematically expressed as \( KL\big(q\left(\mathbf{x}_0 \mid \mathbf{y}\right) \parallel  p\left(\mathbf{x}_0 \mid \mathbf{y}\right)\big) \). Further, under the assumption of zero variance (\( \sigma \approx 0 \)), the optimization problem of seeking \(\mathbf{x}_0\) (\ie, \(\mu\)) can be formulated as minimizing~\cite{song2021maximum, mardani2023variational}:
\vspace{-1.5mm}
\begin{equation}
    \|\mathbf{y}-\mathcal{A}(\mathbf{x}_0)\|^2+w_t(\mathtt{sg}[\epsilon_\phi(\mathbf{x}_t;t)-\epsilon])^\top\mathbf{x}_0,
    \label{eq:RED}
\vspace{-1.5mm}
\end{equation}
where \(w_t\) denotes the loss weights and \(\epsilon\) is sampled from the standard Gaussian distribution.
Here, \( \mathtt{sg} \) signifies the stopped-gradient operator, indicating that backpropagation through the trained diffusion models is not required. The optimization procedure initiates by selecting a timestep \( t \) and applying a perturbation to the target \( \mathbf{x}_0 \) as per Eq.~\eqref{eq:forward diffusion}, resulting in \( \mathbf{x}_t \). Subsequently, the gradients \( [\epsilon_\phi(\mathbf{x}_t;t) - \epsilon] \) are applied to the optimization variable \( \mathbf{x}_0 \). 
In a nutshell, this framework~\cite{mardani2023variational} provides a flexible yet robust strategy for employing diffusion priors in generic optimization problems, serving as a cornerstone for our work.

\begin{figure}[!t]
\centering
\includegraphics[width=\linewidth]{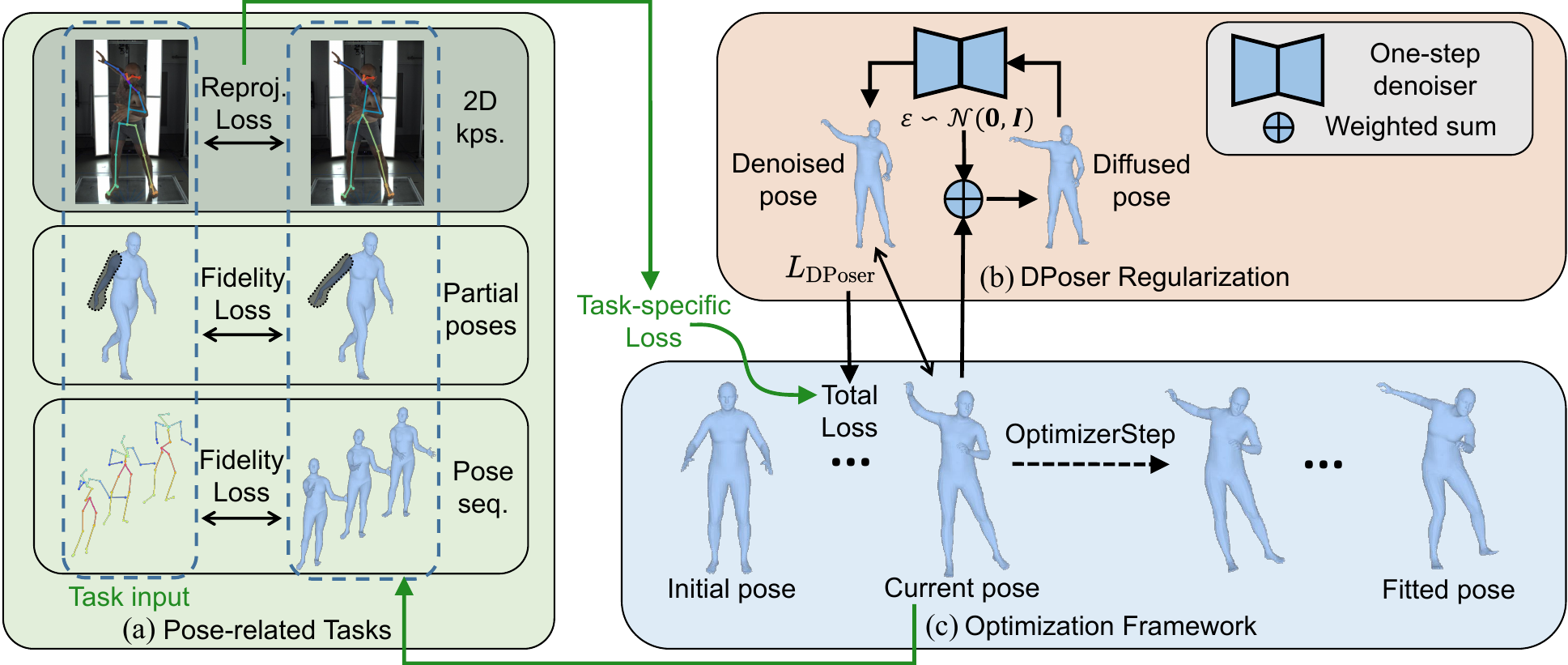}
\vspace{-4.5mm}
\caption{Overview of the DPoser Methodology. Panel (a) presents three tasks: human mesh recovery, pose completion, and motion denoising, with omissions like camera optimization for clarity. Panel (b) demonstrates the DPoser regularization process, introducing noise to the current pose and applying a one-step denoiser to achieve a denoised pose. \(L_\text{DPoser}\) is computed between the denoised and current pose. Panel (c) outlines the optimization process from initial to fitted poses via loss minimization.}
\vspace{-4mm}
\label{fig:overview}
\end{figure}

\noindent \textbf{Introducing DPoser regularization.}\
To shed more light on the working mechanism, we propose to reformulate the regularization term as:
\vspace{-2.5mm}
\begin{align}
    L_\mathrm{DPoser} &= w_t||\mathbf{x}_0-\mathtt{sg}[\mathbf{\hat{x}}_0(t)]||_2^2, \text{where} \label{eq:DPoser}\\
    \mathbf{\hat{x}}_0(t) &= \frac{\mathbf{x}_t - \sigma_{t}\epsilon_\phi(\mathbf{x}_t;t)}{\alpha_{t}}. 
\end{align}
Here, \(\mathbf{\hat{x}}_0(t)\) functions as a precise one-step denoising prediction using the diffusion model \(\epsilon_\phi(\mathbf{x}_t;t)\). This approach effectively encourages the current pose \(\mathbf{x}_0\) towards a denoised, plausible pose distribution, employing a straightforward L2-loss within the DPoser regularization framework. Further, the theoretical foundation of our regularization demonstrates its alignment with the gradient direction of variational diffusion sampling (Eq.~\eqref{eq:RED}).

\emph{Proof: Differentiating Eq.~\eqref{eq:DPoser} with respect to \(\mathbf{x}_0\) yields:}
\begin{align}
\nabla_{\mathbf{x}_0} L_{\mathrm{DPoser}}&=2w_t(\mathbf{x}_0-\mathbf{\hat{x}}_0(t)) \notag\\
&=2w_t(\frac{\mathbf{x}_t - \sigma_{t}\epsilon}{\alpha_{t}} -\frac{\mathbf{x}_t - \sigma_{t}\epsilon_\phi(\mathbf{x}_t;t)}{\alpha_{t}} ) \notag\\
&=2w_t\frac{\sigma_{t}}{\alpha_{t}} (\epsilon_\phi(\mathbf{x}_t;t)-\epsilon) \notag\\
&\propto (\epsilon_\phi(\mathbf{x}_t;t)-\epsilon). 
\label{eq:SDS gradient}
\end{align}

Thus, \(L_\mathrm{DPoser}\) represents a more intuitive approach to variational diffusion sampling. By incorporating alongside task-specific loss functions, this regularization term enhances the plausibility of the resultant poses.

\noindent \textbf{DPoser across pose-related tasks.}\
DPoser excels in versatility, enabling its seamless application in a spectrum of human pose-related tasks. Its adaptability is especially evident in our human mesh recovery approach, as depicted in \cref{fig:overview}.
For an exhaustive examination of DPoser's utility across tasks like pose completion and motion denoising, we direct the reader to our Appendix.

Human mesh recovery aims to deduce the human pose and shape from single-image inputs.
In this context, we refine the optimization function derived from the SMPLify framework~\cite{bogo2016keep}, integrating DPoser as a regularization term, \(L_\mathrm{DPoser}\), and streamlining the process by omitting the intricate interpenetration error component. The modified optimization objective, engaging both pose \(\theta\) and shape \(\beta\) parameters from the SMPL model~\cite{loper2015smpl}, is defined as: 
\begin{equation}
    L(\theta ,\beta ) =  L_J + w_\theta L_\theta + w_\beta L_\beta + w_\alpha L_\mathrm{DPoser}.
    \label{eq:HMR}
\end{equation}
The reprojection loss \(L_J\), acting as the data fidelity measure, is defined by:
\begin{align}
    L_J &= \sum_{i \in \text{Joints}} \lambda_i \rho\left(\Pi_C\left(M_J(\theta, \beta)_i\right) - J^{\text{est}}_i\right),
\end{align}
where \(M_J(\theta, \beta)\) calculates the 3D joint coordinates through SMPL's forward kinematics. The function \(\Pi_C\) maps these 3D coordinates into 2D space, aligning with the camera's perspective. \(J^{\text{est}}\) refers to the 2D keypoints estimated using an off-the-shelf 2D pose estimator (in our case, ViTPose~\cite{xu2022vitpose}), with \(\lambda_i\) reflecting the confidence score for each joint \(i\). The Geman-McClure error function (\(\rho\)) is employed to assess the discrepancy in 2D joint locations reliably.

To mitigate the issue of overfitting, which often leads to unrealistic poses when solely minimizing reprojection loss, several regularization terms are introduced. Specifically, alongside our body prior \(L_\mathrm{DPoser}\), the bending term \(L_\theta\) is incorporated to penalize excessive bending at the elbows and knees, formulated as \(L_\theta = \sum_{i \in \text{(elbows, knees)}} \exp(\boldsymbol{\theta}_i)\). Additionally, the shape regularization term \(L_\beta = \|\beta\|_2^2\) is employed to maintain the body shape within plausible bounds. The weights for prior terms are denoted as \(w_\theta, w_\beta\) and \(w_\alpha\), respectively.

Given the structure of \(L_\mathrm{DPoser}\) (as seen in Eq.~\eqref{eq:DPoser}), a crucial aspect lies in judiciously selecting the diffusion timestep \(t\) during the iterative optimization process. In the subsequent section, we address this concern by introducing our novel truncated timestep scheduling strategy.

\vspace{-2mm}
\subsection{Test-time Truncated Timestep Scheduling}
\noindent \textbf{Motivation from pose generation.}\
Adapting techniques from the image domain to pose data requires a nuanced understanding of the differences between the two. Previous image-based research~\cite{choi2022perception} shows that initial timesteps (larger \(t\)) correspond to the perceptual content, while later timesteps refine details. Pose data, however, lacks this structured layering and spatial redundancy, indicating a need for a tailored timestep approach in the diffusion process.

% In optimizing pose-related tasks, one fundamental challenge arises from the gap between image-based data and articulated pose parameters.
% Prior studies in image domains~\cite{choi2022perception} have noted that earlier timesteps (i.e., larger \(t\)) are typically aligned with perceptual content, while the later timesteps are more closely associated with the refinement of details in the diffusion process. However, this characteristic may not hold for pose data, since they do not have such structured information and much less spatial redundancy.

\begin{wrapfigure}{r}{0.5\textwidth}
  \centering
  \vspace{-5mm}
  \includegraphics[width=0.48\textwidth]{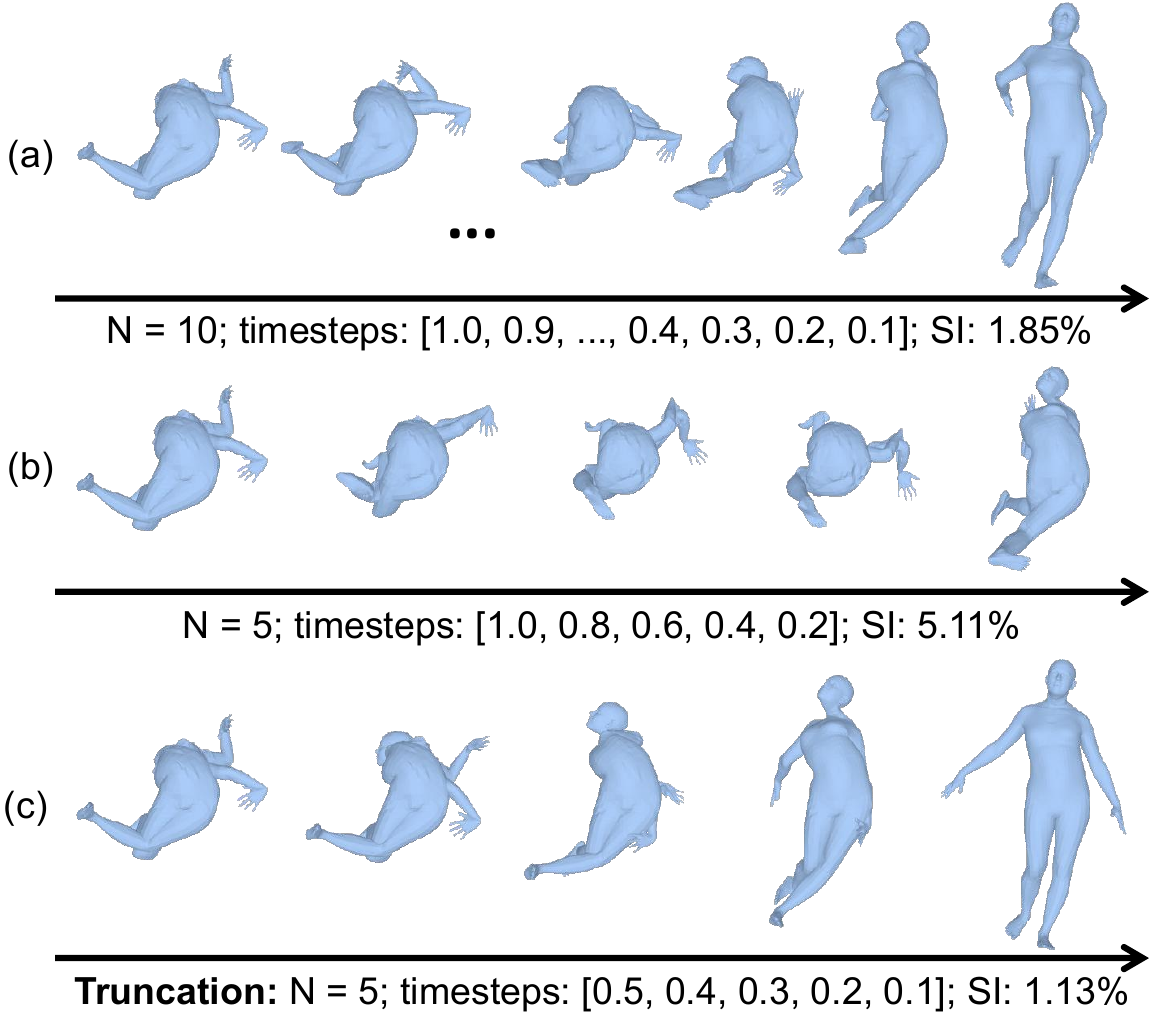}
  \caption{Illustration of the rationale behind our proposed truncated timestep scheduling. We employ the deterministic DDIM sampler~\cite{song2020denoising} with limited steps and assess the quality of generated poses using the Self-Intersection percentage (SI). }
  \label{fig:truncation}
  \vspace{-5mm}
\end{wrapfigure}

As depicted in \cref{fig:truncation}, we find that pose generation doesn't benefit from the early timesteps as image generation does. The significant stages of pose refinement occur at smaller \(t\), specifically when \(t\le0.3\). A uniform distribution of timesteps, as tested in (b) with only five steps, proves less effective for pose data. In contrast, allocating these steps toward the latter end of the diffusion process, as in (c), yields significantly better samples, implying the critical information is not evenly distributed but rather is concentrated toward the end.

\noindent \textbf{Truncated timestep scheduling.}\
Based on these insights, we propose a shift from standard uniform timestep sampling to a truncated strategy, especially for pose data. By focusing on the last timesteps, particularly between 0.2 and 0.0, we target the interval rich in pose-specific information.
Specifically, based on the linear descending scheduling, the truncated timestep \(t\) for each optimization step can be expressed as:
\begin{equation}
     t = t_{\text{max}} - \frac{(t_{\text{max}} - t_{\text{min}}) \times \text{iter}}{N-1}.
\end{equation}
where \(N\) denotes the total number of optimization iterations, and \(\text{iter}\) signifies the current iteration.
This formulation is integral to our proposed optimization framework, which is comprehensively summarized in \cref{alg:test-time-optimization}. The practical implementation typically involves setting the truncated range to \([0.2, 0.05]\).

\begin{algorithm}[!t]
\caption{Test-time Optimization with DPoser}
\label{alg:test-time-optimization}
\begin{algorithmic}[1]
\Require 
A trained diffusion model \( \epsilon_\phi(\mathbf{x}_t; t) \), task-specific loss \( L_{\text{task}} \), range of diffusion timesteps \([t_{\text{max}}, t_{\text{min}}]\), number of optimization iterations \( N \).
\Ensure
Initialization of SMPL body pose parameters \( \mathbf{x}_0 \)
\For{\( \text{iter} = 0, 1, \ldots, N-1 \)}
    \State \( t \leftarrow t_{\text{max}} - \frac{(t_{\text{max}} - t_{\text{min}}) \times \text{iter}}{N-1} \)   \Comment{Timestep scheduling}
    \State Sample \( \epsilon \sim \mathcal{N}(0, I) \)
    \State \( \mathbf{x}_t \leftarrow \alpha_t \mathbf{x}_0 + \sigma_t \epsilon \)  \Comment{Forward diffusion}
    \State \( \mathbf{\hat{x}}_0(t) \leftarrow \frac{\mathbf{x}_t - \sigma_t \epsilon_\phi(\mathbf{x}_t; t)}{\alpha_t} \) \Comment{One-step denoiser}
    \State \( L_{\text{DPoser}} \leftarrow w_t \lVert \mathbf{x}_0 - \text{sg}[\mathbf{\hat{x}}_0(t)] \rVert_2^2 \) \Comment{DPoser regularization}
    \State \( L_{\text{total}} \leftarrow L_{\text{task}} + L_{\text{DPoser}} \)
    \State Update \( \mathbf{x}_0 \) via backpropagation on \( L_{\text{total}} \)
\EndFor
\State \textbf{return} \( \mathbf{x}_0 \)
\end{algorithmic}
\end{algorithm}

\vspace{-1.5mm}
\section{Experiments}
\vspace{-1.5mm}
In this section, we showcase the robustness and versatility of DPoser across a spectrum of pose-centric tasks, including pose generation, human mesh recovery, pose completion, and motion denoising. Due to the page limit, we leave experimental details and more qualitative assessments in the Appendix.

\vspace{-1.5mm}
\subsection{Experimental Setup}
\vspace{-1.5mm}
\noindent \textbf{Implementation details.}\
We train our DPoser model on the AMASS dataset~\cite{mahmood2019amass}, adhering to the same training partition as previous works~\cite{pavlakos2019expressive, tiwari2022pose}.
The model employs axis-angle representation for joint rotations, which we normalize to have zero mean and unit variance. The architecture consists of a fully connected neural network with approximately 8.28M parameters. It draws inspiration from GFPose~\cite{ci2023gfpose} but omits conditional input pathways for our unconditional setting. To stabilize training, we use an exponential moving average with a decay factor of 0.9999, as advised by \cite{song2020score}. The Adam optimizer, a learning rate of \(2 \times 10^{-4}\), and a batch size of 1280 govern the optimization process. 
The training of 800,000 iterations takes roughly 8 hours on a single Nvidia RTX 3090Ti GPU.

\noindent \textbf{Evaluation metrics.}\
To comprehensively evaluate our models across various tasks, following Pose-NDF~\cite{tiwari2022pose}, we adopt task-specific metrics:

\begin{itemize}
\item \textit{Pose Generation}: Diversity and fidelity are evaluated using Average Pairwise Distance (APD) and Self-Intersection rates (SI), respectively.

\item \textit{Human Mesh Recovery}: The Procrustes-aligned Mean Per Joint Position Error (PA-MPJPE) measures the accuracy of recovered human meshes.

\item \textit{Pose Completion}: The Mean Per Joint Position Error (MPJPE) for masked body joints serves as the metric, focusing on the inferred occluded parts.

\item \textit{Motion Denoising}: Both MPJPE and the Mean Per-Vertex Position Error (MPVPE) are calculated to assess the denoising effectiveness.
\end{itemize}
All errors are reported in millimeter units.

% per-vertex

\vspace{-1.5mm}
\subsection{Pose Generation}
\vspace{-1.5mm}
\begin{figure}[t]
    \centering
    \begin{subfigure}[b]{0.32\linewidth}
        \includegraphics[width=\linewidth]{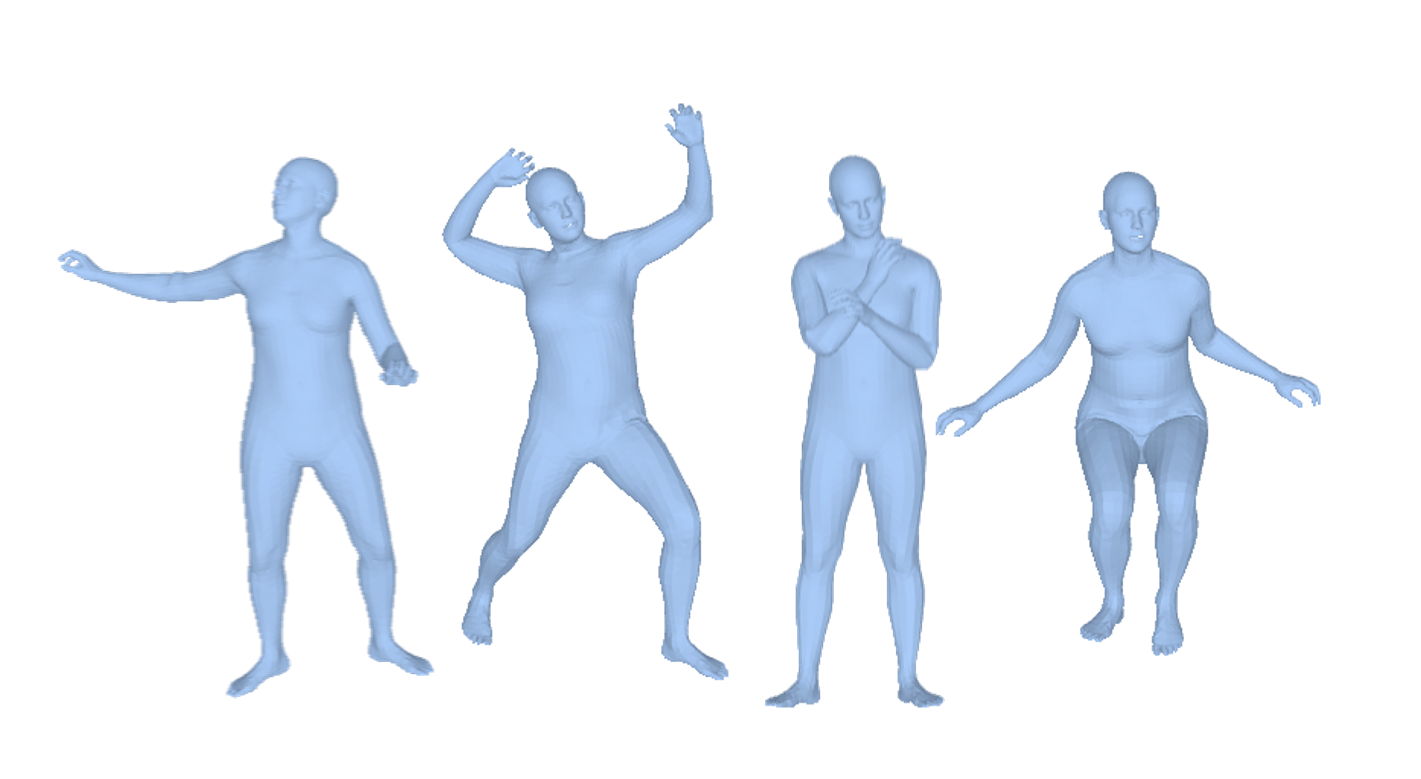}
        \caption{GAN-S~\cite{davydov2022adversarial}}
    \end{subfigure}
    \hfill
    \begin{subfigure}[b]{0.32\linewidth}
        \includegraphics[width=\linewidth]{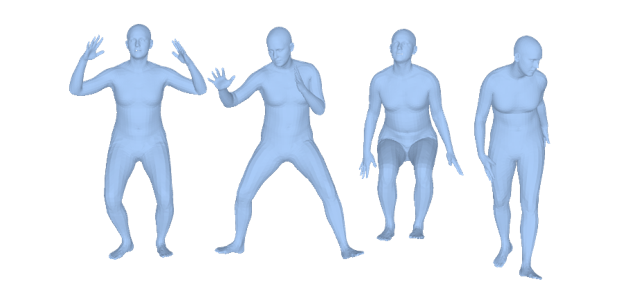}
        \caption{DPoser (ours)}
    \end{subfigure}
    \hfill
    \begin{subfigure}[b]{0.32\linewidth}
        \includegraphics[width=\linewidth]{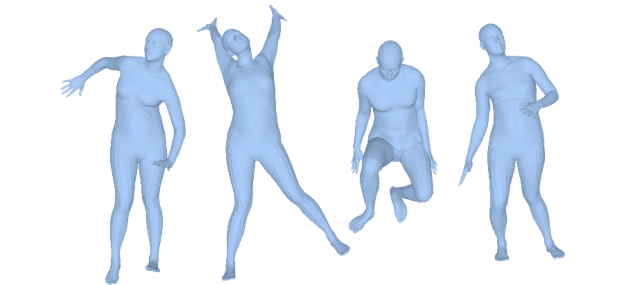}
        \caption{DPoser (ours)*}
    \end{subfigure}
    \hfill
        \begin{subfigure}[b]{0.32\linewidth}
        \includegraphics[width=\linewidth]{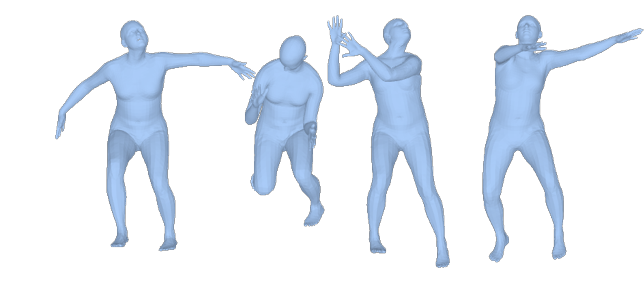}
        \caption{GMM~\cite{bogo2016keep}}
    \end{subfigure}
    \hfill
        \begin{subfigure}[b]{0.32\linewidth}
        \includegraphics[width=\linewidth]{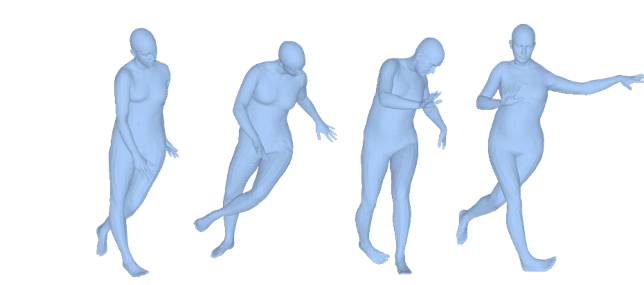}
        \caption{Pose-NDF~\cite{tiwari2022pose}}
    \end{subfigure}
    \hfill
    \begin{subfigure}[b]{0.32\linewidth}
        \includegraphics[width=\linewidth]{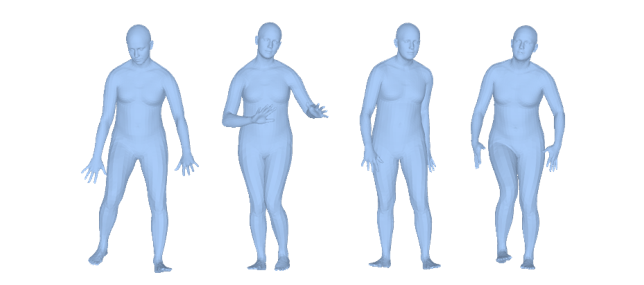}
        \caption{VPoser~\cite{pavlakos2019expressive}}
\end{subfigure}
\caption{Qualitative comparison of generated human poses: (b) illustrates naturalistic poses aligned with real-world data, whereas (c) shows poses that, despite superior metrics, lack natural appearance. *We use a DDIM sampler~\cite{song2020denoising} with only 10 steps.}
\vspace{-6mm}
\label{fig:generation}
\end{figure}

\begin{wraptable}{r}{0.5\textwidth}
  \centering
  \small
  \vspace{-6mm}
  \begin{tabular}{lcc}
    \toprule[\heavyrulewidth]
    Sample source & APD $\uparrow$ & SI $\downarrow$ \\
    \midrule[\lightrulewidth]
    Real-world (AMASS)~\cite{mahmood2019amass} & 15.44 & 0.79 \\
    \midrule[\lightrulewidth]
    GMM~\cite{bogo2016keep} & 16.28 & 1.54 \\
    VPoser~\cite{pavlakos2019expressive} & 10.75 & 1.51 \\
    Pose-NDF~\cite{tiwari2022pose} & 18.75 & 1.97 \\
    GAN-S~\cite{davydov2022adversarial} & 15.68 & 1.27 \\
    \rowcolor{colorTab}
    DPoser (ours) & 14.28 & 1.21 \\
    \rowcolor{colorTab}
    DPoser (ours)* & \textbf{19.03} & \textbf{1.13} \\
    \bottomrule[\heavyrulewidth]
  \end{tabular}
  \caption{Comparative analysis of pose generation metrics. The discrepancy between visual impressions and APD/SI metrics is discussed, with reference to \cref{fig:generation}. *Indicates the use of a reduced 10-step sampler.}
  % *Indicates the use of a DDIM sampler~\cite{song2020denoising} with a reduced 10-step discretization.
  \label{tab:generation}
  \vspace{-8mm}
\end{wraptable}

To commence, we delve into the capabilities of our DPoser model by generating samples from the learned manifold. Employing a standard Euler-Maruyama discretization with 1000 steps, we assess both the diversity and realism of the generated poses (\cref{fig:generation}). While DPoser's outputs are visually diverse and realistic, poses generated from competing methods like GMM~\cite{bogo2016keep} and Pose-NDF~\cite{tiwari2022pose} fall short in naturalism, and VPoser~\cite{pavlakos2019expressive} exhibits limited diversity.
% To commence, we evaluate DPoser's generative abilities using Euler-Maruyama discretization over 1000 steps (\cref{fig:generation}). While DPoser offers visually diverse and realistic poses, competing methods like GMM~\cite{bogo2016keep} and Pose-NDF~\cite{tiwari2022pose} lack naturalism, and VPoser~\cite{pavlakos2019expressive} shows limited diversity.

Interestingly, quantitative metrics such as APD and SI (\cref{tab:generation}) do not always corroborate our qualitative findings. For instance, a 10-step DDIM sampler~\cite{song2020denoising}—suboptimal by design—outperformed real-world data~\cite{mahmood2019amass} in APD, which we attribute to the generation of exaggerated poses.
In summary, our findings underscore the need for a balanced evaluation strategy that merges quantitative metrics with qualitative observations. 

\vspace{-1.5mm}
\subsection{Human Mesh Recovery}
\vspace{-1.5mm}

\begin{table}[t]
  \centering
  \small
  \resizebox{\linewidth}{!}{
    \begin{tabular}{lcccccc}
    \toprule[\heavyrulewidth]
    Initialization & No fitting & GMM~\cite{bogo2016keep} & VPoser~\cite{pavlakos2019expressive} & Pose-NDF~\cite{tiwari2022pose} & GAN-S~\cite{davydov2022adversarial} & DPoser(Ours) \\
    \midrule[\lightrulewidth]
    from scratch  & 108.57 & 58.32 & 58.08 & 57.87 & 57.26 & \textbf{56.05} \\
    CLIFF~\cite{li2022cliff}  & 56.62 & 51.02 & 49.39 & 49.50 & 49.58 & \textbf{49.05} \\
    \bottomrule[\heavyrulewidth]
    \end{tabular}
    }
\vspace{1.5mm}
\caption{Performance comparison of human mesh recovery on the EHF dataset~\cite{pavlakos2019expressive} using two initialization methods. PA-MPJPE is reported as the metric.}
\vspace{-5mm}
\label{tab:HMR}
\end{table}

\begin{figure}
    \centering
    \begin{subfigure}[b]{\linewidth}
        \includegraphics[width=\linewidth]{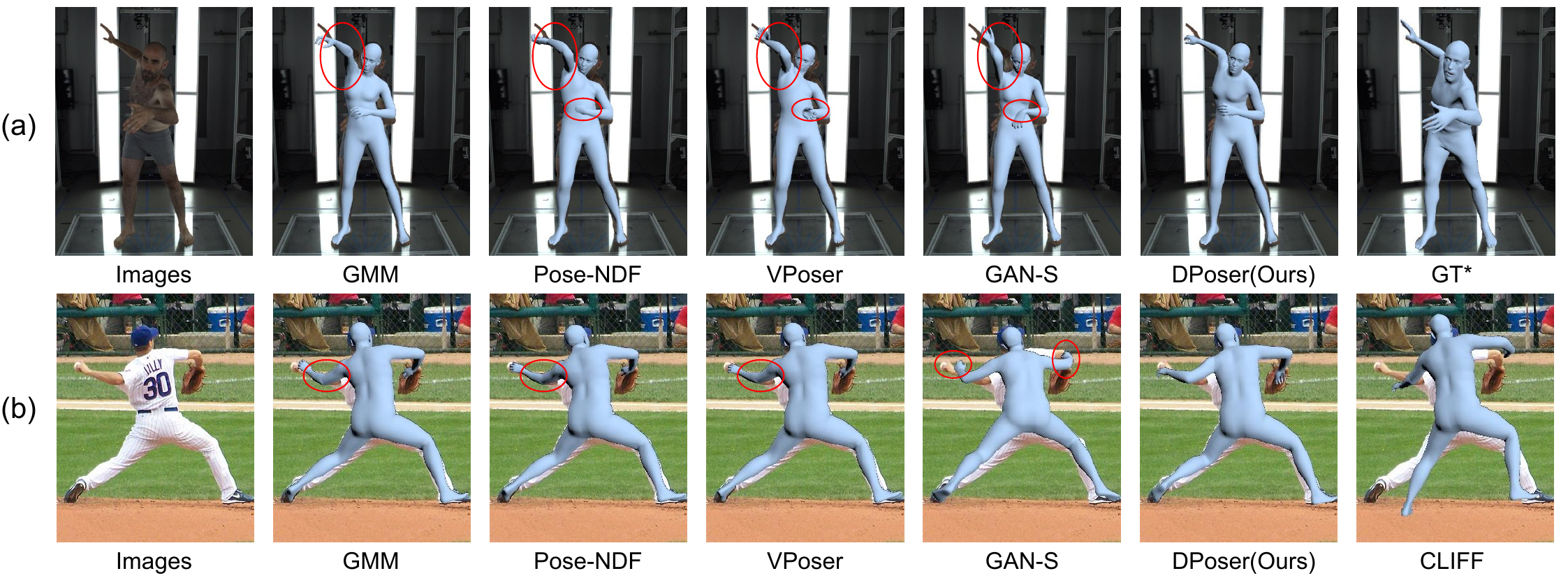}
    \end{subfigure}
    \vspace{-6mm}
    \caption{Human mesh recovery. (a) Fitting from scratch. *Ground truth for the EHF dataset is annotated in SMPL-X~\cite{pavlakos2019expressive}, which extends SMPL~\cite{loper2015smpl} with fully articulated hands and an expressive face. (b) Initialization using the CLIFF~\cite{li2022cliff} prediction.}
    \label{fig:HMR-compare}
    \vspace{-2mm}
\end{figure}

\begin{figure}[t]
    \centering
    \begin{subfigure}[b]{0.49\linewidth}
        \includegraphics[width=\linewidth]{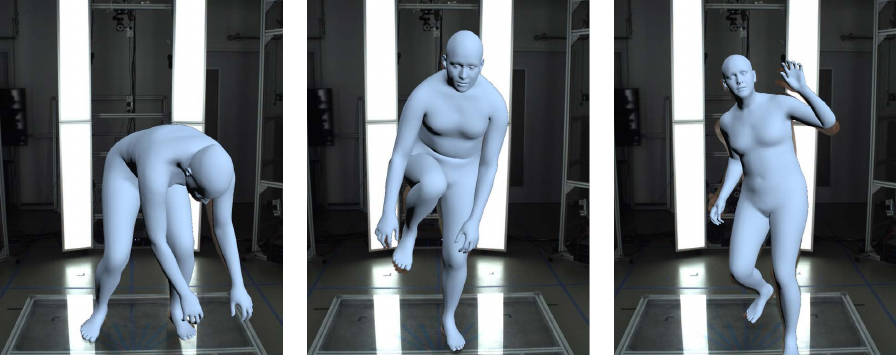}
        \caption{EHF dataset}
    \end{subfigure}
    \hfill
    \begin{subfigure}[b]{0.49\linewidth}
        \includegraphics[width=\linewidth]{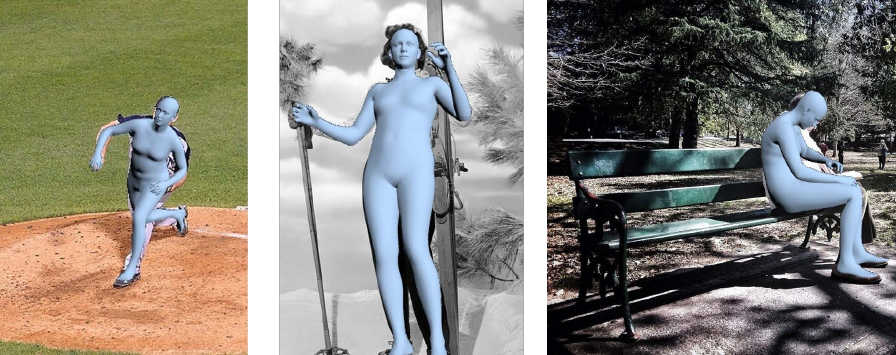}
        \caption{MSCOCO dataset}
    \end{subfigure}
    \hfill
    \begin{subfigure}[b]{0.49\linewidth}
        \includegraphics[width=\linewidth]{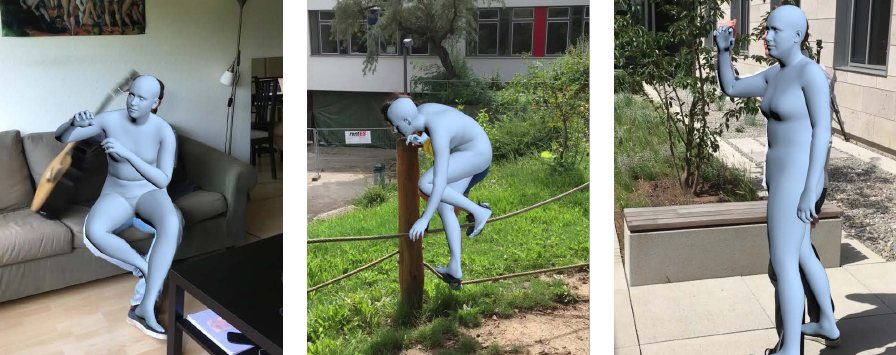}
        \caption{3DPW dataset}
    \end{subfigure}
    \hfill
    \begin{subfigure}[b]{0.49\linewidth}
        \includegraphics[width=\linewidth]{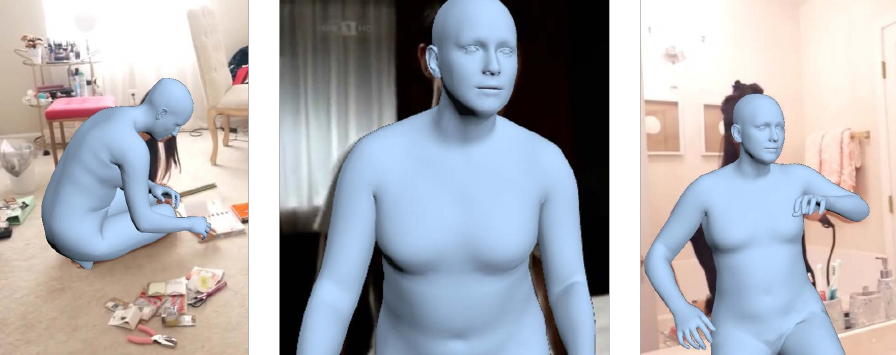}
        \caption{UBody dataset}
    \label{fig:UBody}
    \end{subfigure}
\vspace{-2mm}
\caption{Qualitative evaluations of human mesh recovery leveraging DPoser as pose prior. CLIFF~\cite{li2022cliff} serves as the optimization initializer.}
\vspace{-2mm}
\label{fig:HMR}
\end{figure}

We probe the efficacy of DPoser in human mesh recovery (HMR), focusing on estimating human pose and shape from monocular images. We conduct experiments on the EHF dataset~\cite{pavlakos2019expressive} and benchmark our method against existing SOTA priors. 
Our optimization-based framework incorporates two initialization paradigms: (1) a baseline initialization that utilizes mean pose values and a default camera setup, and (2) an advanced initialization scheme that leverages CLIFF~\cite{li2022cliff}, a pre-trained regression-based model tailored for HMR. Moreover, GAN-S~\cite{davydov2022adversarial} implementations require a GAN-inversion phase to convert initial poses into their latent representations, which is notably time-consuming.

\cref{tab:HMR} and \cref{fig:HMR-compare} showcase the comparative performance of DPoser, highlighting its exceptional ability in HMR tasks. Notably, when fitting from scratch, it surpasses established SOTA priors like GAN-S~\cite{davydov2022adversarial} and Pose-NDF~\cite{tiwari2022pose} and rivals the specific regression-based model~\cite{li2022cliff}. The integration of CLIFF as initialization further amplifies DPoser's performance, underscoring its efficiency and the benefits of employing refined starting conditions. \cref{fig:HMR} further confirms DPoser's superior efficacy and adaptability across multiple datasets including EHF~\cite{pavlakos2019expressive}, MSCOCO~\cite{lin2019microsoft}, 3DPW~\cite{von2018recovering}, and UBody~\cite{lin2023one}.

\vspace{-1.5mm}
\subsection{Pose Completion}
\vspace{-2mm}

In practical scenarios like those encountered in the UBody dataset~\cite{lin2023one} (refer to \cref{fig:UBody}), HMR algorithms often grapple with occlusions leading to incomplete 3D pose estimates. In this context, our ambition is to recover full 3D poses from partially observed data, initializing the occluded parts with random noise. Our DPoser model is employed to refine these initially implausible poses into feasible ones, utilizing an L2 loss on the visible parts to ensure data consistency.

\begin{wraptable}{r}{0.48\textwidth}
      \centering
      \small
      \begin{tabular}{lccc}
        \toprule[\heavyrulewidth]
        Initialization & VPoser & Pose-NDF & DPoser \\
        \midrule[\lightrulewidth]
        Zeros & 180.90 & 157.50 & 73.92 \\
        10mm noise & 181.86 & 172.50 & 74.69 \\
        100mm noise & 180.25 & 511.51 & 74.19 \\
        \bottomrule[\heavyrulewidth]
      \end{tabular}
      \vspace{-2mm}
      \captionof{table}{Pose completion on the AMASS~\cite{mahmood2019amass} dataset (left leg under occlusion, single-hypotheses) using various initialization strategies. DPoser demonstrates its effectiveness across all conditions.}
      \label{tab:completion init}
      \vspace{-6mm}
\end{wraptable}
In parallel, we employ a comparable optimization strategy for both Pose-NDF~\cite{tiwari2022pose} and VPoser~\cite{pavlakos2019expressive}. Notably, \cref{tab:completion init} reveals that Pose-NDF struggles with poorly initialized poses unseen during its training phase. To mitigate this issue, we have to initialize the occluded poses near zero (close to rest pose) for Pose-NDF to prevent optimization divergence. Additionally, as a task-specific baseline, we adapt the original VPoser model into CVPoser by incorporating conditional inputs within its VAE framework~\cite{kingma2013auto}. This modification enables the encoder and decoder to process additional partial poses, facilitating end-to-end conditional sampling.

\begin{table}[t]
\centering
\resizebox{\linewidth}{!}{
    \begin{tabular}{lcccc}
        \toprule[\heavyrulewidth]
        Methods & Occ. left leg & Occ. legs & Occ. arms & Occ. trunk\\
        \midrule[\lightrulewidth]
        PoseNDF (\(S=1\))~\cite{tiwari2022pose} & 158.21 & 159.19 & 201.00 & 75.42 \\
        PoseNDF (\(S=5\)) & 147.66/158.11/7.62 & 151.86/159.21/5.33 & 196.36/200.92/3.30 & 70.88/75.39/3.25 \\
        PoseNDF (\(S=10\)) & 144.38/158.06/8.31 & 149.38/159.14/5.90 & 194.79/200.87/3.63 & 69.45/75.38/3.54 \\
        VPoser (S=1)~\cite{pavlakos2019expressive} & 180.78 & 198.18 & 159.86 & 37.75 \\
        VPoser (\(S=5\)) & 167.92/181.30/10.53 & 178.77/198.15/14.51 & 148.17/159.65/8.64 & 31.83/37.79/4.54 \\
        VPoser (\(S=10\)) & 162.82/181.09/12.21 & 172.83/198.31/16.30 & 144.53/159.80/9.69 & 30.06/37.78/4.99 \\
        \midrule[\lightrulewidth]
        CVPoser (\(S=10\)) \(^\dagger\) & 71.66/145.52/51.68 & 90.49/148.30/38.46 & 83.02/136.82/36.47 & 18.77/37.83/13.12 \\
        \midrule[\lightrulewidth]
        \rowcolor{colorTab}
        DPoser(ours) (\(S=1\)) & \textbf{74.48} & \textbf{97.39} & \textbf{81.49} & \textbf{28.58} \\
        \rowcolor{colorTab}
        DPoser(ours) (\(S=5\)) & \textbf{42.64}/73.85/24.36 & \textbf{67.70}/97.06/22.29 & \textbf{58.52}/82.37/18.33 & \textbf{17.11}/28.59/8.92 \\
        \rowcolor{colorTab}
        DPoser(ours) (\(S=10\)) & \textbf{35.37}/74.01/26.47 & \textbf{59.25}/96.77/24.55 & \textbf{51.27}/81.76/20.04 & \textbf{13.95}/28.57/9.85 \\
        \bottomrule[\heavyrulewidth]
    \end{tabular}
    }
\vspace{1.5mm}
\caption{Performance metrics (min/mean/std of MPJPE across multiple hypotheses) for pose completion under varying occlusion scenarios. \(S\) denotes the number of hypotheses. \(^\dagger\) Task-specific baseline trained with partial poses as conditional input.}
\vspace{-6mm}
\label{tab:completion}
\end{table}

\begin{figure}[t]
\centering
\includegraphics[width=0.98\linewidth]{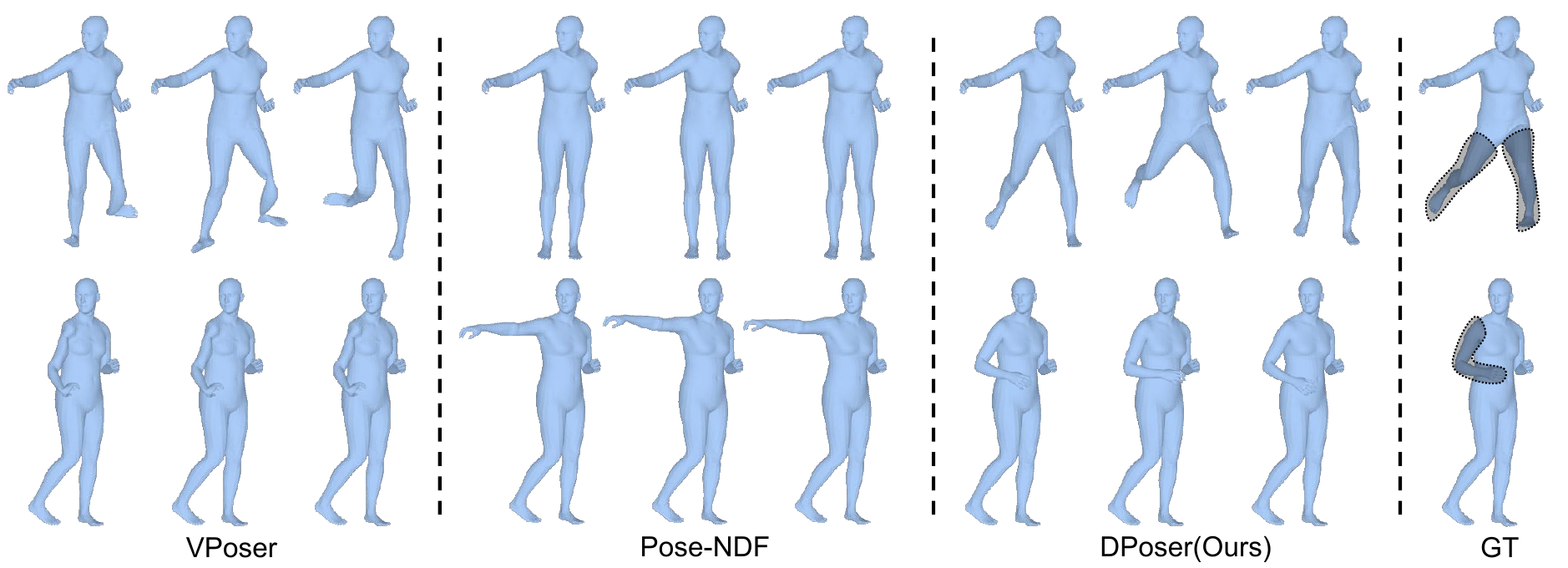}
\vspace{-2mm}
\caption{Qualitative evaluations of pose completion. Three hypotheses are drawn for each method. DPoser uniquely offers multiple plausible solutions for partially observed poses, a scenario where competitors often struggle due to limited generalization. 
}
\vspace{-2mm}
\label{fig:completion}
\end{figure}

Given the inherent uncertainties within this task, we generate multiple solutions and evaluate them based on their minimum, mean, and standard deviation errors against the ground truth.
As illustrated in \cref{tab:completion}, DPoser exhibits superior performance across different occlusion scenarios compared to existing pose priors and even the task-specific CVPoser, highlighting its effectiveness in pose completion. 
The qualitative evaluations are presented in \cref{fig:completion}. Here, we observe that DPoser can generate a multitude of plausible poses, a capability lacking in VPoser~\cite{pavlakos2019expressive}. Pose-NDF~\cite{tiwari2022pose}, meanwhile, struggles with generalizing to unseen noisy poses and making plausible adjustments from its rest pose initialization.

\vspace{-2mm}
\subsection{Motion Denoising}
\vspace{-1.5mm}
Though not initially designed for temporal tasks, DPoser shows remarkable proficiency in motion denoising. The task aims to estimate clean body poses from noisy 3D joint positions in motion capture sequences.
Adhering to the setup outlined in HuMoR~\cite{rempe2021humor}, we utilize 60-frame sequences from the AMASS~\cite{mahmood2019amass} dataset and artificially introduce Gaussian noise with a standard deviation of 40 mm to the 3D joint positions. Moreover, we conduct experiments on HPS datasets~\cite{guzov2021human} without additional training to validate the generalization. 

As presented in \cref{tab:motion}, DPoser sets a new standard in motion denoising, outperforming even specialized motion priors like HuMoR~\cite{rempe2021humor}. 
To further confirm the robustness of DPoser, we conduct evaluations under varying conditions to gauge DPoser's denoising capabilities. The results, detailed in \cref{tab:motion2}, reveal that DPoser consistently achieves significant reductions in MPJPE, maintaining robust performance under extreme noise conditions. 
% , specifically at a noise standard deviation of 100.00 mm

\vspace{-1.5mm}
\subsection{Ablation Study}
\vspace{-1.5mm}

\begin{figure}[t]
    \centering
    \begin{minipage}{.5\textwidth}
        \centering
        \small
        \begin{tabular}{lcc}
        \toprule[\heavyrulewidth]
        Methods & AMASS~\cite{mahmood2019amass} & HPS~\cite{guzov2021human}\\
            \midrule[\lightrulewidth]
            No prior & 24.19 & 23.67 \\
            VPoser~\cite{pavlakos2019expressive} & 23.42 & 22.78 \\
            Pose-NDF~\cite{tiwari2022pose} & 22.13 & 21.60 \\
            MVAE~\cite{ling2020character} & 26.80 & N/A \\
            HuMoR~\cite{rempe2021humor} & 22.69 & N/A \\
            \midrule[\lightrulewidth]
            \rowcolor{colorTab}
            DPoser (ours) & \textbf{19.87} & \textbf{20.54} \\
            \bottomrule[\heavyrulewidth]
        \end{tabular}
        \vspace{-2mm}
        \captionof{table}{Performance metrics (MPJPE) for motion denoising under 40 mm noise.}
        \label{tab:motion}       
    \end{minipage}\hfill
    \begin{minipage}{.45\textwidth}
        \centering
          \centering
          \small
          \begin{tabular}{lcc}
            \toprule[\heavyrulewidth]
            Noise std & AMASS~\cite{mahmood2019amass} & HPS~\cite{guzov2021human} \\
            \midrule[\lightrulewidth]
            20.00 & 31.93/13.64 & 31.93/13.45 \\
            40.00 &  63.81/19.87 & 63.81/20.54 \\
            100.00 & 159.78/33.18 & 159.78/35.32 \\
            \bottomrule[\heavyrulewidth]
          \end{tabular}
          \vspace{-2mm}
          \captionof{table}{DPoser in motion denoising under varying noise scales. MPJPE is reported as before/after applying DPoser denoising.}
          \label{tab:motion2}    
    \end{minipage}
    \vspace{-2mm}
\end{figure}

\begin{table}[t]
\centering
\resizebox{\linewidth}{!}{
\setlength{\tabcolsep}{2mm}
\begin{tabular}{lcccc}
\toprule[\heavyrulewidth]
\makecell[c]{\multirow{2}{*}[-1.0ex]{Timestep scheduling}}& \multicolumn{1}{c}{HMR} & \multicolumn{1}{c}{Pose Completion} & \multicolumn{2}{c}{Motion Denoising} \\
    \cmidrule(r){2-2} \cmidrule(lr){3-3} \cmidrule(l){4-5}
    & \makecell[c]{PA-MPJPE $\downarrow$} & \makecell[c]{MPJPE (\(S=10\)) $\downarrow$} & \makecell[c]{MPVPE $\downarrow$} & \makecell[c]{MPJPE $\downarrow$} \\
    \midrule[\lightrulewidth]
    Random & 58.84 & 86.23/121.57/23.16 & 43.33 & 23.87 \\
    Fixed & 56.55 & 36.99/71.68/23.41 & 45.69 & 22.54 \\
    Uniform & 59.28 & 42.72/75.70/21.84 & 39.72 & 20.80 \\
    \rowcolor{colorTab}
    Truncated & \textbf{56.05} & \textbf{35.37}/74.01/26.47 & \textbf{38.21} & \textbf{19.87} \\
    \bottomrule[\heavyrulewidth]
\end{tabular}
}
\vspace{0.5mm}
\caption{Evaluation of timestep scheduling strategies on key pose-related tasks, highlighting the superior efficacy of the proposed truncated scheduling.}
\vspace{-5mm}
\label{tab:scheduling}
\end{table}

In our ablation study, we initially focus on the impact of truncated timestep scheduling on DPoser's performance. This involves contrasting our proposed scheduling strategy against three established methods—random, fixed, and uniform scheduling~\cite{muller2023generative, mardani2023variational, chung2022diffusion, song2020score}.
As \cref{tab:scheduling} demonstrates, our strategy consistently outperforms these alternatives across all evaluated tasks. 
Additionally, we delve into the training aspects of DPoser, such as rotation representations and the integration of an auxiliary loss akin to HuMoR~\cite{rempe2021humor}. 
Using the same trained prior, we also compare DPoser's capabilities with SOTA diffusion-based solvers~\cite{song2020score, chung2022improving, chung2022diffusion} on pose completion, revealing its superior versatility and performance.
Detailed findings and analyses from these ablation studies are presented in the Appendix.

\vspace{-2.5mm}
\section{Conclusion}
\vspace{-1.5mm}
We introduce DPoser, to our best knowledge, the first unconditional diffusion-based pose prior, tailored for an expansive array of pose-related tasks. Engineered for flexibility, DPoser can be implemented as a straightforward L2-loss regularizer and enhanced by our innovative truncated timestep scheduling for test-time optimization. 
Comprehensive experiments substantiate DPoser's superior performance over existing state-of-the-art pose priors.

\noindent \textbf{Limitation and future work.}\
While our framework benefits from variational diffusion sampling~\cite{mardani2023variational}, it also shares its limitations, such as the mode-seeking behavior. 
Future research could look into enhancing solution diversity via techniques like particle-based variational inference~\cite{liu2016stein, wang2023prolificdreamer}.
Furthermore, within the broader context of inverse problems we have framed, a plethora of existing methods~\cite{song2022pseudoinverse, boys2023tweedie, chung2023parallel, murata2023gibbsddrm} could be adapted to leverage our diffusion-based prior. Exploring these methods holds great potential for future progress.

\noindent \textbf{Ethical Considerations.}
For a discussion on the potential negative impacts of our work, please refer to the Appendix.

\bibliographystyle{splncs04}
\bibliography{main}

\begin{thebibliography}{10}
\providecommand{\url}[1]{\texttt{#1}}
\providecommand{\urlprefix}{URL }
\providecommand{\doi}[1]{https://doi.org/#1}

\bibitem{bogo2016keep}
Bogo, F., Kanazawa, A., Lassner, C., Gehler, P., Romero, J., Black, M.J.: Keep it smpl: Automatic estimation of 3d human pose and shape from a single image. In: ECCV (2016)

\bibitem{boys2023tweedie}
Boys, B., Girolami, M., Pidstrigach, J., Reich, S., Mosca, A., Akyildiz, O.D.: Tweedie moment projected diffusions for inverse problems. arXiv preprint arXiv:2310.06721  (2023)

\bibitem{cho2023generative}
Cho, H., Kim, J.: Generative approach for probabilistic human mesh recovery using diffusion models. In: ICCV (2023)

\bibitem{choi2022perception}
Choi, J., Lee, J., Shin, C., Kim, S., Kim, H., Yoon, S.: Perception prioritized training of diffusion models. In: CVPR (2022)

\bibitem{chung2023parallel}
Chung, H., Kim, J., Kim, S., Ye, J.C.: Parallel diffusion models of operator and image for blind inverse problems. In: CVPR (2023)

\bibitem{chung2022diffusion}
Chung, H., Kim, J., Mccann, M.T., Klasky, M.L., Ye, J.C.: Diffusion posterior sampling for general noisy inverse problems. arXiv preprint arXiv:2209.14687  (2022)

\bibitem{chung2022improving}
Chung, H., Sim, B., Ryu, D., Ye, J.C.: Improving diffusion models for inverse problems using manifold constraints. NeurIPS  (2022)

\bibitem{ci2023gfpose}
Ci, H., Wu, M., Zhu, W., Ma, X., Dong, H., Zhong, F., Wang, Y.: Gfpose: Learning 3d human pose prior with gradient fields. In: CVPR (2023)

\bibitem{davydov2022adversarial}
Davydov, A., Remizova, A., Constantin, V., Honari, S., Salzmann, M., Fua, P.: Adversarial parametric pose prior. In: CVPR (2022)

\bibitem{deng2009imagenet}
Deng, J., Dong, W., Socher, R., Li, L.J., Li, K., Fei-Fei, L.: Imagenet: A large-scale hierarchical image database. In: 2009 IEEE conference on computer vision and pattern recognition. pp. 248--255. Ieee (2009)

\bibitem{dhariwal2021diffusion}
Dhariwal, P., Nichol, A.: Diffusion models beat gans on image synthesis. NeurIPS  (2021)

\bibitem{georgakis2020hierarchical}
Georgakis, G., Li, R., Karanam, S., Chen, T., Ko{\v{s}}eck{\'a}, J., Wu, Z.: Hierarchical kinematic human mesh recovery. In: ECCV (2020)

\bibitem{creswell2018generative}
Goodfellow, I., Pouget-Abadie, J., Mirza, M., Xu, B., Warde-Farley, D., Ozair, S., Courville, A., Bengio, Y.: Generative adversarial networks. Communications of the ACM  (2020)

\bibitem{graikos2022diffusion}
Graikos, A., Malkin, N., Jojic, N., Samaras, D.: Diffusion models as plug-and-play priors. NeurIPS  (2022)

\bibitem{guzov2021human}
Guzov, V., Mir, A., Sattler, T., Pons-Moll, G.: Human poseitioning system (hps): 3d human pose estimation and self-localization in large scenes from body-mounted sensors. In: CVPR (2021)

\bibitem{ho2020denoising}
Ho, J., Jain, A., Abbeel, P.: Denoising diffusion probabilistic models. NeurIPS  (2020)

\bibitem{holmquist2023diffpose}
Holmquist, K., Wandt, B.: Diffpose: Multi-hypothesis human pose estimation using diffusion models. In: ICCV (2023)

\bibitem{jiang2023back}
Jiang, Z., Zhou, Z., Li, L., Chai, W., Yang, C.Y., Hwang, J.N.: Back to optimization: Diffusion-based zero-shot 3d human pose estimation. arXiv preprint arXiv:2307.03833  (2023)

\bibitem{kanazawa2018end}
Kanazawa, A., Black, M.J., Jacobs, D.W., Malik, J.: End-to-end recovery of human shape and pose. In: CVPR (2018)

\bibitem{karras2022elucidating}
Karras, T., Aittala, M., Aila, T., Laine, S.: Elucidating the design space of diffusion-based generative models. NeurIPS  (2022)

\bibitem{kawar2022denoising}
Kawar, B., Elad, M., Ermon, S., Song, J.: Denoising diffusion restoration models. NeurIPS  (2022)

\bibitem{kingma2013auto}
Kingma, D.P., Welling, M.: Auto-encoding variational bayes. arXiv preprint arXiv:1312.6114  (2013)

\bibitem{kipf2016semi}
Kipf, T.N., Welling, M.: Semi-supervised classification with graph convolutional networks. arXiv preprint arXiv:1609.02907  (2016)

\bibitem{li2023diffhand}
Li, L., Zhuo, L., Zhang, B., Bo, L., Chen, C.: Diffhand: End-to-end hand mesh reconstruction via diffusion models. arXiv preprint arXiv:2305.13705  (2023)

\bibitem{li2022cliff}
Li, Z., Liu, J., Zhang, Z., Xu, S., Yan, Y.: Cliff: Carrying location information in full frames into human pose and shape estimation. In: ECCV (2022)

\bibitem{lin2023one}
Lin, J., Zeng, A., Wang, H., Zhang, L., Li, Y.: One-stage 3d whole-body mesh recovery with component aware transformer. In: CVPR (2023)

\bibitem{lin2019microsoft}
Lin, T.Y., Maire, M., Belongie, S., Bourdev, L., Girshick, R., Hays, J., Perona, P., Ramanan, D., Zitnick, C.L., Doll{\'a}r, P.: Microsoft coco: common objects in context (2014). arXiv preprint arXiv:1405.0312  (2019)

\bibitem{ling2020character}
Ling, H.Y., Zinno, F., Cheng, G., Van De~Panne, M.: Character controllers using motion vaes. TOG  (2020)

\bibitem{liu2016kernelized}
Liu, Q., Lee, J., Jordan, M.: A kernelized stein discrepancy for goodness-of-fit tests. In: ICML (2016)

\bibitem{liu2016stein}
Liu, Q., Wang, D.: Stein variational gradient descent: A general purpose bayesian inference algorithm. NeurIPS  (2016)

\bibitem{loper2015smpl}
Loper, M., Mahmood, N., Romero, J., Pons-Moll, G., Black, M.J.: Smpl: A skinned multi-person linear model. ACM Transactions on Graphics  \textbf{34}(6) (2015)

\bibitem{mahmood2019amass}
Mahmood, N., Ghorbani, N., Troje, N.F., Pons-Moll, G., Black, M.J.: Amass: Archive of motion capture as surface shapes. In: ICCV (2019)

\bibitem{mardani2023variational}
Mardani, M., Song, J., Kautz, J., Vahdat, A.: A variational perspective on solving inverse problems with diffusion models. arXiv preprint arXiv:2305.04391  (2023)

\bibitem{muller2023generative}
M{\"u}ller, L., Ye, V., Pavlakos, G., Black, M., Kanazawa, A.: Generative proxemics: A prior for 3d social interaction from images. arXiv preprint arXiv:2306.09337  (2023)

\bibitem{murata2023gibbsddrm}
Murata, N., Saito, K., Lai, C.H., Takida, Y., Uesaka, T., Mitsufuji, Y., Ermon, S.: Gibbsddrm: A partially collapsed gibbs sampler for solving blind inverse problems with denoising diffusion restoration. arXiv preprint arXiv:2301.12686  (2023)

\bibitem{nachmani2021non}
Nachmani, E., Roman, R.S., Wolf, L.: Non gaussian denoising diffusion models. arXiv preprint arXiv:2106.07582  (2021)

\bibitem{pavlakos2019expressive}
Pavlakos, G., Choutas, V., Ghorbani, N., Bolkart, T., Osman, A.A., Tzionas, D., Black, M.J.: Expressive body capture: 3d hands, face, and body from a single image. In: CVPR (2019)

\bibitem{poole2022dreamfusion}
Poole, B., Jain, A., Barron, J.T., Mildenhall, B.: Dreamfusion: Text-to-3d using 2d diffusion. arXiv preprint arXiv:2209.14988  (2022)

\bibitem{qiu2023learning}
Qiu, Z., Yang, Q., Wang, J., Wang, X., Xu, C., Fu, D., Yao, K., Han, J., Ding, E., Wang, J.: Learning structure-guided diffusion model for 2d human pose estimation. arXiv preprint arXiv:2306.17074  (2023)

\bibitem{rempe2021humor}
Rempe, D., Birdal, T., Hertzmann, A., Yang, J., Sridhar, S., Guibas, L.J.: Humor: 3d human motion model for robust pose estimation. In: ICCV (2021)

\bibitem{sarkka2019applied}
S{\"a}rkk{\"a}, S., Solin, A.: Applied stochastic differential equations, vol.~10. Cambridge University Press (2019)

\bibitem{shafir2023human}
Shafir, Y., Tevet, G., Kapon, R., Bermano, A.H.: Human motion diffusion as a generative prior. arXiv preprint arXiv:2303.01418  (2023)

\bibitem{sohl2015deep}
Sohl-Dickstein, J., Weiss, E., Maheswaranathan, N., Ganguli, S.: Deep unsupervised learning using nonequilibrium thermodynamics. In: ICML (2015)

\bibitem{song2020denoising}
Song, J., Meng, C., Ermon, S.: Denoising diffusion implicit models. arXiv preprint arXiv:2010.02502  (2020)

\bibitem{song2022pseudoinverse}
Song, J., Vahdat, A., Mardani, M., Kautz, J.: Pseudoinverse-guided diffusion models for inverse problems. In: ICLR (2022)

\bibitem{song2021maximum}
Song, Y., Durkan, C., Murray, I., Ermon, S.: Maximum likelihood training of score-based diffusion models. NeurIPS  (2021)

\bibitem{song2019generative}
Song, Y., Ermon, S.: Generative modeling by estimating gradients of the data distribution. NeurIPS  (2019)

\bibitem{song2020score}
Song, Y., Sohl-Dickstein, J., Kingma, D.P., Kumar, A., Ermon, S., Poole, B.: Score-based generative modeling through stochastic differential equations. arXiv preprint arXiv:2011.13456  (2020)

\bibitem{tiwari2022pose}
Tiwari, G., Anti{\'c}, D., Lenssen, J.E., Sarafianos, N., Tung, T., Pons-Moll, G.: Pose-ndf: Modeling human pose manifolds with neural distance fields. In: ECCV (2022)

\bibitem{vincent2011connection}
Vincent, P.: A connection between score matching and denoising autoencoders. Neural computation  (2011)

\bibitem{von2018recovering}
Von~Marcard, T., Henschel, R., Black, M.J., Rosenhahn, B., Pons-Moll, G.: Recovering accurate 3d human pose in the wild using imus and a moving camera. In: ECCV (2018)

\bibitem{wang2023score}
Wang, H., Du, X., Li, J., Yeh, R.A., Shakhnarovich, G.: Score jacobian chaining: Lifting pretrained 2d diffusion models for 3d generation. In: CVPR (2023)

\bibitem{wang2023prolificdreamer}
Wang, Z., Lu, C., Wang, Y., Bao, F., Li, C., Su, H., Zhu, J.: Prolificdreamer: High-fidelity and diverse text-to-3d generation with variational score distillation. arXiv preprint arXiv:2305.16213  (2023)

\bibitem{wu2023hd}
Wu, J., Gao, X., Liu, X., Shen, Z., Zhao, C., Feng, H., Liu, J., Ding, E.: Hd-fusion: Detailed text-to-3d generation leveraging multiple noise estimation. arXiv preprint arXiv:2307.16183  (2023)

\bibitem{xu2022vitpose}
Xu, Y., Zhang, J., Zhang, Q., Tao, D.: Vi{TP}ose: Simple vision transformer baselines for human pose estimation. In: Advances in Neural Information Processing Systems (2022)

\bibitem{zhao2023modiff}
Zhao, M., Liu, M., Ren, B., Dai, S., Sebe, N.: Modiff: Action-conditioned 3d motion generation with denoising diffusion probabilistic models. arXiv preprint arXiv:2301.03949  (2023)

\bibitem{zhou2019continuity}
Zhou, Y., Barnes, C., Lu, J., Yang, J., Li, H.: On the continuity of rotation representations in neural networks. In: CVPR (2019)

\bibitem{zhu2023hifa}
Zhu, J., Zhuang, P.: Hifa: High-fidelity text-to-3d with advanced diffusion guidance. arXiv preprint arXiv:2305.18766  (2023)

\end{thebibliography}

\setcounter{section}{0}
\setcounter{table}{0}
\setcounter{figure}{0}

\renewcommand{\thesection}{\Alph{section}}   
\renewcommand {\thetable} {S-\arabic{table}}
\renewcommand {\thefigure} {S-\arabic{figure}}

\pagebreak

\begin{center}
    \textbf{\Large Appendix for DPoser: Diffusion Model as \\ Robust 3D Human Pose Prior}
    \vspace{0.8cm}
\end{center}

In this appendix, we first briefly recap the parameterization of diffusion models and their connection to score functions in \cref{sec:detail_intro}, followed by the perspective of Score Distillation Sampling (SDS) to understand our DPoser regularization in \cref{sec:sds}. We detail the experimental setup and nuances in \cref{sec:exp} and dissect various training aspects of DPoser in \cref{sec:training}. The exploration of extended optimization techniques is discussed in \cref{sec:testing}, and considerations for truncated timestep scheduling in image domains are presented in \cref{sec:image_domain}. Additional qualitative results are showcased in \cref{sec:visual}. Lastly, potential negative impacts such as biases in data and ethical concerns in application are considered in \cref{sec:ethical}.

\section{Parameterization of Score-based Diffusion Models}
\label{sec:detail_intro}
In the seminal work by Song \etal~\cite{song2020score}, it is demonstrated that both score-based generative models~\cite{song2019generative} and diffusion probabilistic models~\cite{ho2020denoising} can be understood as discretized versions of stochastic differential equations (SDEs) defined by score functions. This unification allows the training objective to be interpreted either as learning a time-dependent denoiser or as learning a sequence of score functions that describe increasingly noisy versions of the data.

We begin by revisiting the training objective for score-based models~\cite{song2019generative} to elucidate the link with diffusion models~\cite{ho2020denoising}. Consider the transition kernel of the forward diffusion process \(p_{0t}(\mathbf{x}_t | \mathbf{x}_0)=\mathcal{N}(\mathbf{x}_t;\alpha_{t}\mathbf{x}_0,\sigma_{t}^2\mathbf{I})\). Our goal is to learn score functions \( \nabla_{\mathbf{x}_t} \log p_t(\mathbf{x}_t) \) through a neural network \( s_\theta(\mathbf{x}_t; t) \), by minimizing the L2 loss as follows (we omit the expectation operator for conciseness) :
\begin{equation}
    \mathbb{E} \left[w(t)||s_\theta(\mathbf{x}_t;t)-\nabla_{\mathbf{x}_t}\log p_t\left(\mathbf{x}_t\right)||_2^2\right].
    \label{eq:ESM}
\end{equation}
Here, \(\mathbf{x}_t = \alpha_{t}\mathbf{x}_0+\sigma_{t}\epsilon\), where \(\epsilon \sim \mathcal{N}(\mathbf{0}, \mathbf{I})\).

Based on denoising score matching~\cite{vincent2011connection}, we know the minimizing objective Eq.~\eqref{eq:ESM} is equivalent to the following tractable term:
\begin{equation}
    \mathbb{E} \left[w(t)||s_\theta(\mathbf{x}_t;t)-\nabla_{\mathbf{x}_t}\log p_{0t}(\mathbf{x}_t | \mathbf{x}_0)||_2^2\right].
    \label{eq:DSM}
\end{equation}
To link this with the noise predictor \( \epsilon_\theta(\mathbf{x}_t; t) \) in diffusion models, we can employ the reparameterization \( s_\theta(\mathbf{x}_t; t) = - \frac{\epsilon_\theta(\mathbf{x}_t; t)}{\sigma_t} \). Then, Eq.~\eqref{eq:DSM} can be simplified as follows:
\begin{align}
&w(t)||- \frac{\epsilon_\theta(\mathbf{x}_t;t)}{\sigma_t} - \nabla_{\mathbf{x}_t}\log p_{0t}(\mathbf{x}_t \mid \mathbf{x}_0)||_2^2 \notag\\
=&w(t)||- \frac{\epsilon_\theta(\mathbf{x}_t;t)}{\sigma_t} + \frac{(\mathbf{x}_t-\alpha _t\mathbf{x}_0)}{\sigma_{t}^2} ||_2^2 \notag\\
=&w(t)||- \frac{\epsilon_\theta(\mathbf{x}_t;t)}{\sigma_t} + \frac{\sigma_{t}\epsilon}{\sigma_{t}^2})||_2^2 \notag\\
=&\frac{w(t)}{\sigma_{t}^2} ||\epsilon_\theta(\mathbf{x}_t;t) - \epsilon)||_2^2 \label{eq:DSM2}
\end{align}

The resulting form of Eq.~\eqref{eq:DSM2} aligns precisely with the noise prediction form of diffusion models~\cite{ho2020denoising} (refer to Eq.~(4) in the main text). This implies that by training \( \epsilon_\theta(\mathbf{x}_t; t) \) in a diffusion model context, we simultaneously get a handle on the score function, approximated as \( \nabla_{\mathbf{x}_t} \log p_t(\mathbf{x}_t) \approx - \frac{\epsilon_\theta(\mathbf{x}_t; t)}{\sigma_t} \).

\section{View DPoser as Score Distillation Sampling}
\label{sec:sds}
\begin{table}[!t]
  \centering
  \setlength\tabcolsep{6pt}
  \begin{tabular}{lcccc}
    \toprule[\heavyrulewidth]
    \makecell[c]{\multirow{2}{*}[-1.5ex]{Strategy}} & \multicolumn{1}{c}{HMR} & \multicolumn{1}{c}{Pose Completion} & \multicolumn{2}{c}{Motion Denoising} \\
    \cmidrule(r){2-2} \cmidrule(lr){3-3} \cmidrule(l){4-5}
    & \makecell[c]{PA-MPJPE $\downarrow$} & \makecell[c]{MPJPE (\(S=10\)) $\downarrow$} & \makecell[c]{MPVPE $\downarrow$} & \makecell[c]{MPJPE $\downarrow$} \\
    \midrule[\lightrulewidth]
    1 step & \textbf{56.05} & \textbf{35.37}/74.01/26.47 & \textbf{38.21} & \textbf{19.87} \\
    5 steps & 56.16 & 36.59/80.82/31.22 & 40.22 & 21.21 \\
    10 steps & 56.18 & 36.78/82.59/32.32 & 40.69 & 21.34 \\
    \bottomrule[\heavyrulewidth]
  \end{tabular}
  \vspace{2mm}
  \caption{Efficacy of different denoising steps in DPoser's optimization.}
  \label{tab:denoiser steps}
  \vspace{-4mm}
\end{table}

Interestingly, the gradient of DPoser (Eq.~(10) in the main text) coincides with Score Distillation Sampling (SDS)~\cite{poole2022dreamfusion, wang2023score}, which can be interpreted as aiming to minimize the following KL divergence:
\begin{equation}
    KL\big(p_{0t}\left(\mathbf{x}_t\mid \mathbf{x}_0\right) \parallel p_t^\mathtt{SDE}\left(\mathbf{x}_t;\theta \right)\big),
    \label{eq:SDS KL}
\end{equation}
where \(p_t^\mathtt{SDE}\left(\mathbf{x}_t;\theta\right)\) denote the marginal distribution whose score function is estimated by \(\epsilon_\theta(\mathbf{x}_t;t)\).
For the specific case where \(t \to 0\), this term encourages the Dirac distribution \(\delta (\mathbf{x}_0)\) (\ie, the optimized variable) to gravitate toward the learned data distribution \(p_0^\mathtt{SDE}\left(\mathbf{x}_0;\theta\right)\), while the Gaussian perturbation like Eq.~\eqref{eq:SDS KL} softens the constraint.
Building on this understanding, we can borrow advanced techniques from SDS~\cite{poole2022dreamfusion, wang2023score}—a rapidly evolving area ripe for methodological innovations~\cite{wang2023prolificdreamer,wu2023hd,zhu2023hifa}. To extend this, we experiment with a multi-step denoising strategy adapted from HiFA~\cite{zhu2023hifa}, substituting our original one-step denoising process. This alternative, however, yields suboptimal results across all evaluation metrics, as demonstrated in \cref{tab:denoiser steps}. A plausible explanation could be that our proposed truncated timestep scheduling effectively manages low noise levels (\ie, small \(t\)), thus negating the need for more denoising steps. In addition, iterative denoising in each optimization step may cause error accumulations, leading to inaccurate gradients.

\section{Experimental Details}
\label{sec:exp}
This section elaborates on the specifics of our pose completion and motion denoising experiments.

\subsection{Pose Completion}
For partial observations \(\mathbf{y}\), the measurement operator \(\mathcal{A}\) is modeled as a mask matrix \(M \in \mathbb{R}^{d\times n}\). Based on our optimization framework (Algorithm~1 in the main text), we define the task-specific loss, \(L_\text{comp}\), as follows:
\begin{equation}
    L_\text{comp} = ||M \mathbf{x}_0 - \mathbf{y}||_2^2. 
\end{equation}
Here, \(\mathbf{x}_0\) denotes the complete body pose \(\theta\) we try to recover, where the unseen parts are initialized as random noise. In the following ablated studies, if not specified, the evaluation is performed using 10 hypotheses on the AMASS~\cite{mahmood2019amass} dataset with left leg occlusion.
\subsection{Motion Denoising (Noisy Input)}
\label{sec:motion_denoising}
\begin{table}[!t]
\centering
\setlength\tabcolsep{6pt}
\begin{tabular}{lcccc}
    \toprule[\heavyrulewidth]
    \makecell[c]{\multirow{2}{*}[-1.0ex]{Methods}} & \multicolumn{2}{c}{AMASS~\cite{mahmood2019amass}} & \multicolumn{2}{c}{HPS~\cite{guzov2021human}}\\
    \cmidrule(r){2-3} \cmidrule(l){4-5}
    & \makecell[c]{20mm} & \makecell[c]{100mm} & \makecell[c]{20mm} & \makecell[c]{100mm} \\
    \midrule[\lightrulewidth]
    No prior & 15.33 & 51.48 & 16.26 & 50.87 \\
    VPoser~\cite{pavlakos2019expressive} & 15.20 & 49.10 & 17.24 & 46.69 \\
    Pose-NDF~\cite{tiwari2022pose} & 13.84 & 46.10 & 15.62 & 47.50  \\
    \rowcolor{colorTab}
    DPoser (ours) & \textbf{13.64} & \textbf{33.18} & \textbf{13.45} & \textbf{35.32} \\
    \bottomrule[\heavyrulewidth]
  \end{tabular}
  \vspace{2mm}
  \caption{Performance comparison of motion denoising under varying noise scales. MPJPE is reported afters denoising.}
  \label{tab:more motion}
  \vspace{-4mm}
\end{table}

Adhering to Pose-NDF settings~\cite{tiwari2022pose}, we aim to refine noisy joint positions \(J_{\text{obs}}^t\) over \(N\) frames to obtain clean poses \(\theta^t\),  initialized from mean poses in SMPL with small noise. We formulate the task-specific loss combining an observation fidelity term \(L_\text{obs}\) and a temporal consistency term \(L_\text{temp}\):
\begin{equation}
    L_\text{obs} = \sum_{t=0}^{N-1}  ||M_J(\theta^t, \beta_0) - J_{\text{obs}}^t||_2^2,
    \label{eq:motion}
\end{equation}
\begin{equation}
    L_\text{temp} = \sum_{t=1}^{N-1}  ||M_J(\theta^{t-1}, \beta_0) - M_J(\theta^t, \beta_0)||_2^2,
\end{equation}
where \(M_J\) denotes the 3D joint positions regressed from SMPL~\cite{loper2015smpl} and \(\beta_0\) is the constant mean shape parameters.

In complement to the comparative analysis presented in Table 4 of our main text, we extend our evaluation to include scenarios with varying noise levels. This extended examination, detailed in \cref{tab:more motion}, showcases DPoser's exceptional performance against state-of-the-art (SOTA) pose priors, especially under conditions of high noise, manifesting DPoser's resilience to noise.

\subsection{Motion Denoising (Partial Input)}
This task focuses on reconstructing clean poses, \(\theta^t\), from partially observed joint positions, \(J_{\text{obs}}^t\), across \(N\) frames, employing a known mask matrix to identify visible joints. The optimization objective mirrors that of motion denoising (\cref{sec:motion_denoising}), but incorporates a mask in Eq.~\eqref{eq:motion} to specifically target visible parts, ensuring that only these segments guide the recovery process.

We conducted experiments on the AMASS dataset~\cite{mahmood2019amass} to assess our model's performance on this task with two types of occlusions: legs and left arm. The quantitative results of these experiments are detailed in \cref{tab:motion_completion}, and the accompanying visualizations are provided in \cref{sec:visual}.

In leg occlusion scenarios, the AMASS dataset primarily showcases straight poses, offering minimal diversity. This scenario permits decent outcomes without incorporating a pose prior, since the optimization's starting point closely aligns with these prevalent poses. However, VPoser's mean-centered characteristic hinders its ability to faithfully replicate the visible areas. On the other hand, Pose-NDF falls short in enhancing the occluded parts. DPoser accurately handles visible parts and guides occluded ones for more realistic poses. For left arm occlusions, which involve more varied movements, DPoser markedly surpasses other methods, underlining its adaptability and precision in handling diverse motion patterns. 

\begin{table*}[!t]
  \centering
  \setlength\tabcolsep{6pt}
  \begin{tabular}{llcccc}
    \toprule[\heavyrulewidth]
    \makecell[c]{\multirow{2}{*}[-1.0ex]{Methods}} & \makecell[c]{\multirow{2}{*}[-1.0ex]{Occlusion}} & \multicolumn{3}{c}{MPJPE} & MPVPE \\
    \cmidrule(lr){3-5} \cmidrule(l){6-6}
    & & Vis. & Occ. & All. & All. \\
    \midrule[\lightrulewidth]
    No prior & Legs & 0.26 & 14.72 & 5.52 & 5.45 \\
    VPoser & Legs & 1.75 & 14.29 & 6.31 & 7.38 \\
    PoseNDF & Legs & \textbf{0.25} & 15.71 & 5.87 & 5.64 \\
    \rowcolor{colorTab}
    DPoser (ours) & Legs & 0.28 & \textbf{12.24} & \textbf{4.63} & \textbf{3.65} \\
    \midrule[\lightrulewidth]
    No prior & Left Arm & 0.26 & 24.87 & 4.74 & 9.91 \\
    VPoser & Left Arm & 1.21 & 13.23 & 3.40 & 7.68 \\
    PoseNDF & Left Arm & \textbf{0.25} & 17.70 & 3.42 & 7.86 \\
    \rowcolor{colorTab}
    DPoser (ours) & Left Arm & 0.27 & \textbf{7.80} & \textbf{1.64} & \textbf{3.81} \\
    \bottomrule[\heavyrulewidth]
  \end{tabular}
  \vspace{2mm}
  \caption{Comparative analysis of methods for motion denoising with different occlusions (Legs and Left Arm) on the AMASS dataset. Errors (in \emph{cm}) are evaluated in terms of MPJPE across visible (Vis.), occluded (Occ.), and all joints, along with MPVPE for all vertices.}
  \label{tab:motion_completion}
  \vspace{-2mm}
\end{table*}

\section{Ablated DPoser's Training}
\label{sec:training}
\begin{table*}[!t]
  \centering
  \setlength\tabcolsep{5pt}
  \begin{tabular}{lcccccc}
    \toprule[\heavyrulewidth]
    \makecell[c]{\multirow{2}{*}[-1.5ex]{Normalization}} & \multicolumn{1}{c}{HMR} & \multicolumn{1}{c}{Pose Completion} & \multicolumn{2}{c}{Motion Denoising} \\
    \cmidrule(r){2-2} \cmidrule(lr){3-3} \cmidrule(l){4-5}
    & \makecell[c]{PA-MPJPE $\downarrow$} & \makecell[c]{MPJPE (\(S=10\)) $\downarrow$} & \makecell[c]{MPVPE $\downarrow$} & \makecell[c]{MPJPE $\downarrow$} \\
    \midrule[\lightrulewidth]
    w/o norm & 57.88 & 45.37/102.28/41.08 & 44.82 & 24.04 \\
    min-max & 59.17 & 47.41/107.00/43.42 & 42.70 & 21.29 \\
    z-score & \textbf{56.49} & \textbf{34.37}/72.47/26.32 & \textbf{38.57} & \textbf{20.24} \\ 
    \bottomrule[\heavyrulewidth]
  \end{tabular}
  \vspace{2mm}
  \caption{Evaluation of DPoser's performance under different normalization methods, specifically for the axis-angle rotation representation.}
  \label{tab:normalize}
  \vspace{-4mm}
\end{table*}

\begin{table*}[!t]
  \centering
  \setlength\tabcolsep{5pt}
  \begin{tabular}{lcccccc}
    \toprule[\heavyrulewidth]
    \makecell[c]{\multirow{2}{*}[-1.5ex]{Representation}} & \multicolumn{1}{c}{HMR} & \multicolumn{1}{c}{Pose Completion} & \multicolumn{2}{c}{Motion Denoising} \\
    \cmidrule(r){2-2} \cmidrule(lr){3-3} \cmidrule(l){4-5}
    & \makecell[c]{PA-MPJPE $\downarrow$} & \makecell[c]{MPJPE (\(S=10\)) $\downarrow$} & \makecell[c]{MPVPE $\downarrow$} & \makecell[c]{MPJPE $\downarrow$} \\
    \midrule[\lightrulewidth]
    axis-angle & \textbf{56.05} & \textbf{34.76}/72.41/26.09 & \textbf{38.21} & \textbf{19.87} \\
    6D rotations & 57.54 & 40.89/81.43/27.31 & 38.44 & 20.12 \\
    \bottomrule[\heavyrulewidth]
  \end{tabular}
  \vspace{2mm}
  \caption{Comparative performance of rotation representations under z-score normalization across multiple tasks and metrics.}
  \label{tab:rotation}
  \vspace{-4mm}
\end{table*}

This section dissects the impact of different rotation representations and normalization techniques on DPoser's performance. Initially, we examine axis-angle representation, comparing various normalization strategies: min-max scaling, z-score normalization, and no normalization. Our findings, summarized in \cref{tab:normalize}, indicate that z-score normalization is generally the most effective. Subsequently, using this optimal normalization, we explore 6D rotations~\cite{zhou2019continuity} as an alternative. As evidenced by \cref{tab:rotation}, axis-angle representation offers superior performance. This preference can be attributed to the effective modeling capabilities of diffusion models, along with the inherent advantages of axis-angle in capturing bounded joint rotations for regression tasks like human mesh recovery.

Inspired by HuMoR~\cite{rempe2021humor}, we experiment with integrating the SMPL body model~\cite{loper2015smpl} as a regularization term during training. Alongside the prediction of additive noise, as outlined in Equation (4) in the main text, we employ a 10-step DDIM sampler~\cite{song2020denoising} to recover a ``clean'' version of the pose, denoted as \(\tilde{\mathbf{x}}_0\), from the diffused \(\mathbf{x}_t\). The regularization loss aims to minimize the discrepancy between the original and recovered poses under the SMPL body model \(M\):
\begin{equation}
L_\mathrm{reg} =  ||M_J(\tilde{\mathbf{x}}_0, \beta_0) - M_J(\mathbf{x}_0, \beta_0)||_2^2  + ||M_V(\tilde{\mathbf{x}}_0, \beta_0) - M_V(\mathbf{x}_0, \beta_0)||_2^2 .
\end{equation}
Here, \(\beta_0\) represents the mean shape parameters in SMPL. To account for denoising errors, we scale the regularization loss by \(\mathrm{log} (1+\frac{\alpha_t}{\sigma_t})\), thereby increasing the weight for samples with smaller \(t\) values (less noise). 

\begin{wrapfigure}{r}{0.6\textwidth}
    \centering
    \vspace{-2mm}
    \includegraphics[width=\linewidth]{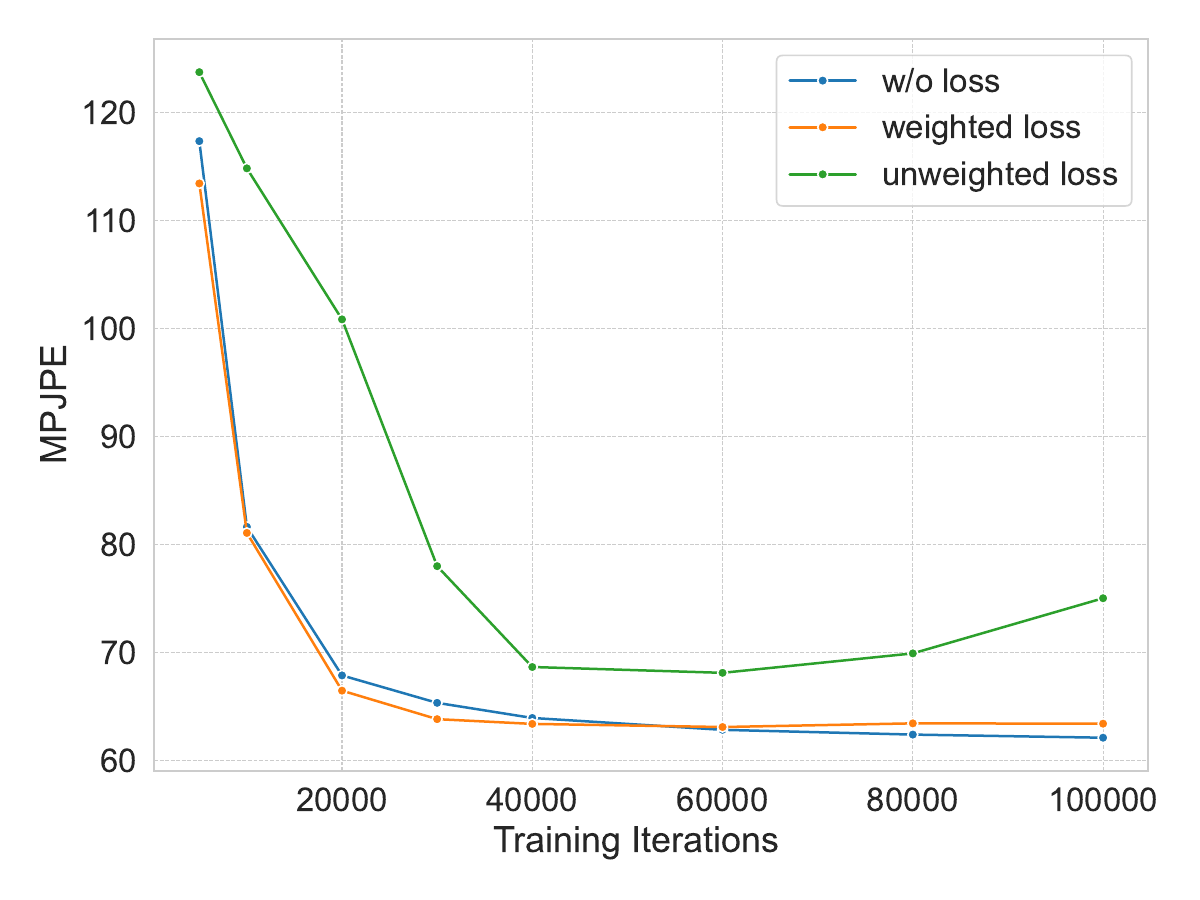}
    \caption{MPJPE evolution in DPoser training for pose completion, assessed on AMASS~\cite{mahmood2019amass} with 10 hypotheses under legs occlusion scenarios.}
    \label{fig:regloss}
    \vspace{-10mm}
\end{wrapfigure}

\cref{fig:regloss} visualizes the impact of this regularization on MPJPE during the training, specifically for pose completion tasks with occlusion of both legs. 
We observe that weighted regularization offers slight performance gains in the early training process, while the absence of weighting introduces instability and deterioration in results. Despite these insights, the computational cost of incorporating the SMPL model—especially for our large batch size of 1280—makes the training approximately 8 times slower. Therefore, we opted not to include this regularization in our main experiments.

\section{Extended DPoser's Optimization}
\label{sec:testing}
\begin{table}[!t]
  \centering
  \resizebox{\linewidth}{!}{
    \begin{tabular}{lcccc}
    \toprule[\heavyrulewidth]
    Methods & Occ. left leg & Occ. legs & Occ. arms & Occ. trunk\\
    \midrule[\lightrulewidth]
    ScoreSDE~\cite{song2020score} & 48.73/106.32/41.30 & 74.68/128.32/37.27 & 66.89/127.86/48.15 & 16.69/34.54/12.21 \\
    DPS~\cite{chung2022diffusion} & 40.51/104.32/54.57 & 64.26/113.46/33.71 & 60.63/119.85/42.78 & 15.10/33.90/13.27 \\
    MCG~\cite{chung2022improving} & 49.04/106.37/41.07 & 74.90/128.53/37.40 & 66.17/127.72/48.15 & 16.69/34.66/12.23 \\
    \rowcolor{colorTab}
    DPoser(ours) & \textbf{35.37}/74.01/26.47 & \textbf{59.25}/96.77/24.55 & \textbf{51.27}/81.76/20.04 & \textbf{13.95}/28.57/9.85 \\
    \bottomrule[\heavyrulewidth]
    \end{tabular}
    }
    \vspace{2mm}
  \caption{Comparative evaluation of diffusion-based solvers for pose completion on the AMASS dataset~\cite{mahmood2019amass} (hypotheses number \(S=10\)).}
  \label{tab:solvers}
  \vspace{-4mm}
\end{table}

In addressing pose-centric tasks as inverse problems, we propose a versatile optimization framework, which employs variational diffusion sampling as its foundational approach~\cite{mardani2023variational}. Our exploration extends to an array of diffusion-based methodologies for solving these complex inverse problems. Among the techniques considered are ScoreSDE~\cite{song2020score}, MCG~\cite{chung2022improving}, and DPS~\cite{chung2022diffusion}.
These methods augment standard generative processes with observational data, either by employing gradient-based guidance or back-projection techniques.
We compare these methods with our DPoser for pose completion tasks. Our findings, captured in \cref{tab:solvers}, reveal that DPoser outperforms the competitors under most occlusion conditions. Consequently, DPoser emerges not merely as a universally applicable solution to pose-related tasks, but also as an exceptionally efficient one.

It is worth mentioning that methods rooted in generative frameworks~\cite{song2020score, chung2022improving, chung2022diffusion, kawar2022denoising} can pose challenges for broader applicability in pose-centric tasks. For instance, in blind inverse problems—certain parameters in \(\mathcal{A}\) (e.g., camera models in HMR) are unknown—generative methods are less straightforward to implement. 
ZeDO~\cite{jiang2023back}, a recent study focusing on the 2D-3D lifting task, adopts the ScoreSDE~\cite{song2020score} framework and refines camera translations by solving an optimization sub-problem after each generative step.
However, directly porting this strategy to HMR is non-trivial, owing to the added complexity of body shape parameter optimization—a feature currently absent in our DPoser model. 
Although some state-of-the-art techniques~\cite{chung2023parallel, murata2023gibbsddrm} offer solutions by jointly modeling operator \(\mathcal{A}\) and data distributions, a full-fledged discussion on this subject is beyond this paper's purview and remains an open question for future work.

\section{Truncated Timestep Scheduling on Images}
\label{sec:image_domain}
\begin{figure*}[t]
    \centering
    \includegraphics[width=\linewidth]{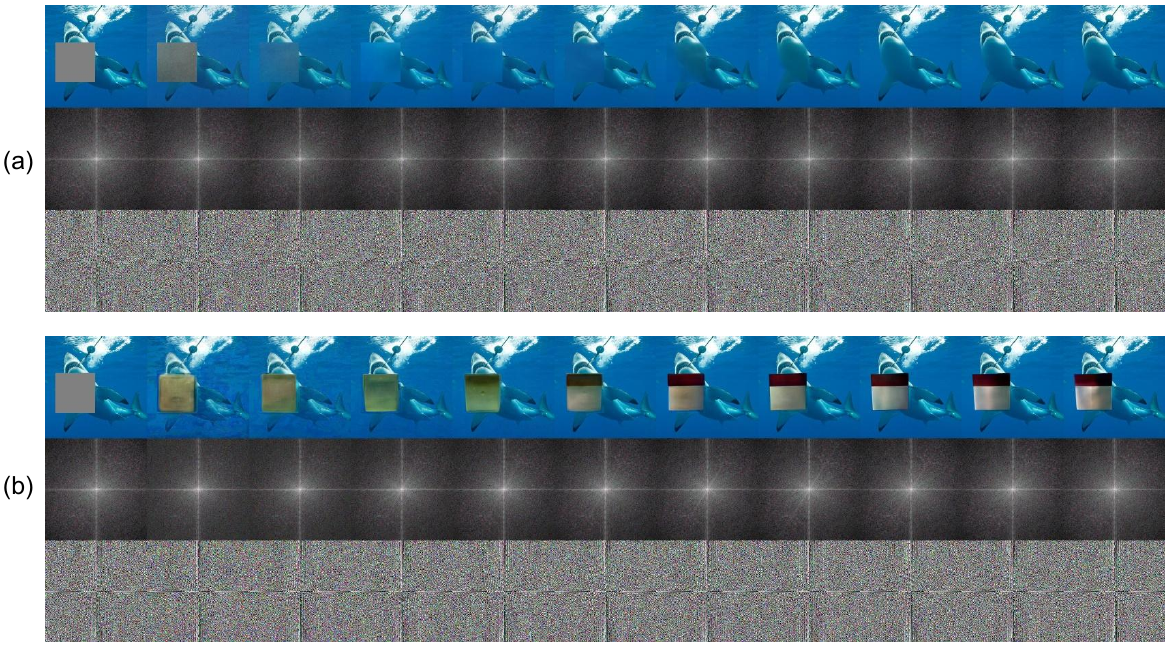}
    \caption{Image inpainting using standard (a) and truncated (b) timestep scheduling. The process evolution is shown over iterations with the middle row depicting the log-magnitude spectrum and the bottom row the phase spectrum.}
     % (a) exhibits cohesive restoration with detail fidelity; (b) results in detail-rich patches that are perceptually incongruent with the original image context.
    \label{fig:truncated_images}
\end{figure*}
Exploring truncated timestep scheduling for image-based tasks, we find its suitability for human poses doesn't translate well to images. Initial timesteps are critical in image domains for generating foundational perceptual content.

In our study, we employed a 256x256 unconditional diffusion model~\cite{dhariwal2021diffusion} trained on ImageNet~\cite{deng2009imagenet} with variational diffusion sampling~\cite{mardani2023variational} for image inpainting. Comparing standard (timesteps 990 to 0) and truncated scheduling (timesteps 495 to 0), both with 100 steps, the experiments confirmed that truncation compromises image quality (\cref{fig:truncated_images}). The standard approach preserved perceptual content, while truncation produced disjointed patches, misaligned with the original image context.

These results affirm that truncated timestep scheduling excels in pose data where key information emerges in later stages but falls short in image tasks where early timesteps are essential. This scheduling is thus bespoke to the characteristics of human pose estimation and is unsuitable for image processes that rely on the full diffusion timeline for content fidelity.

\section{More Qualitative Results}
\label{sec:visual}
\begin{figure*}
    \centering
    \includegraphics[width=\linewidth]{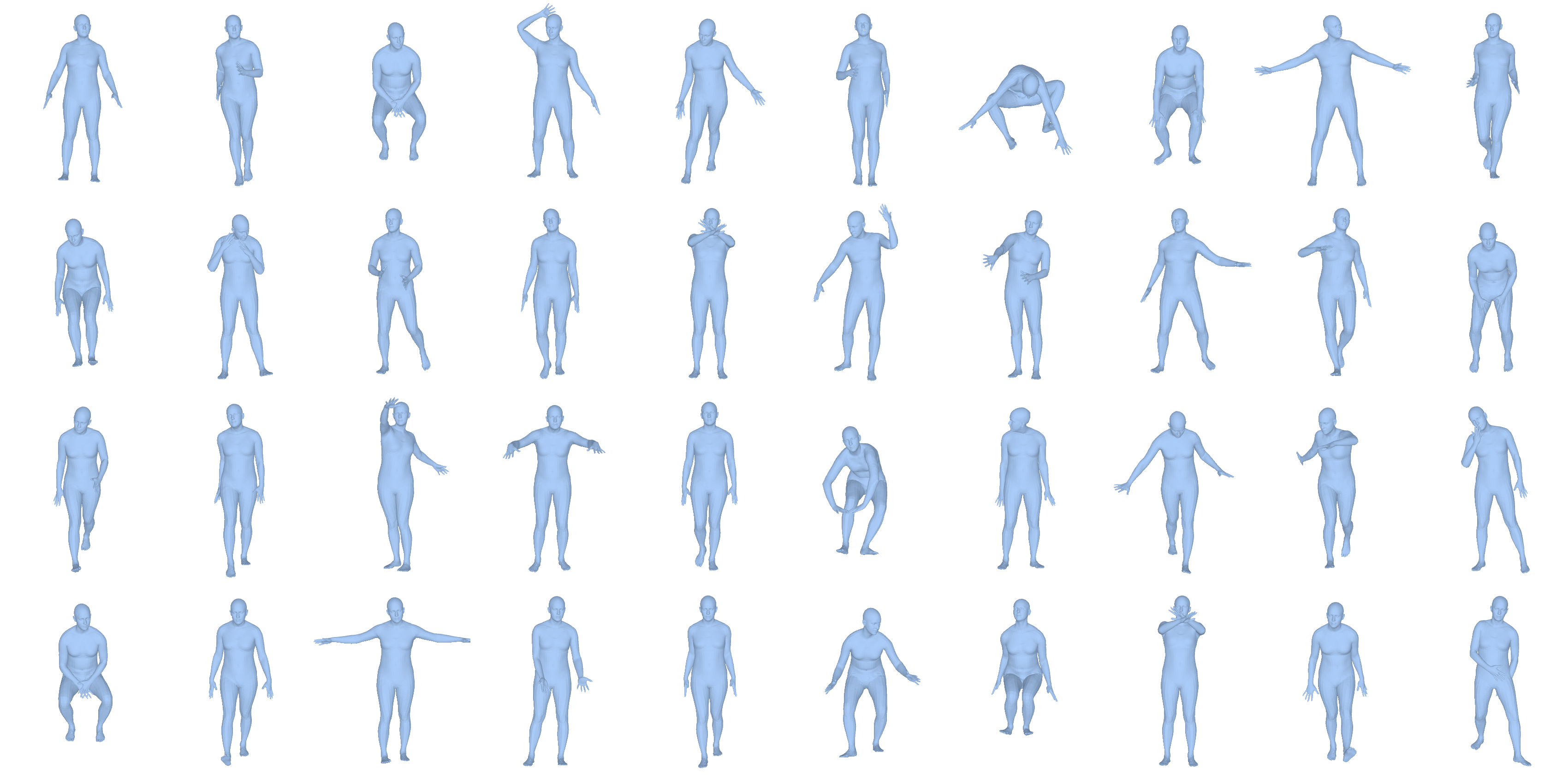}
    \vspace{-6mm}
    \caption{Pose generation. DPoser can generate diverse and realistic poses.}
    \label{fig:more generation}
\end{figure*}

\begin{figure*}
    \centering
    \begin{subfigure}[b]{\linewidth}
        \includegraphics[width=\linewidth]{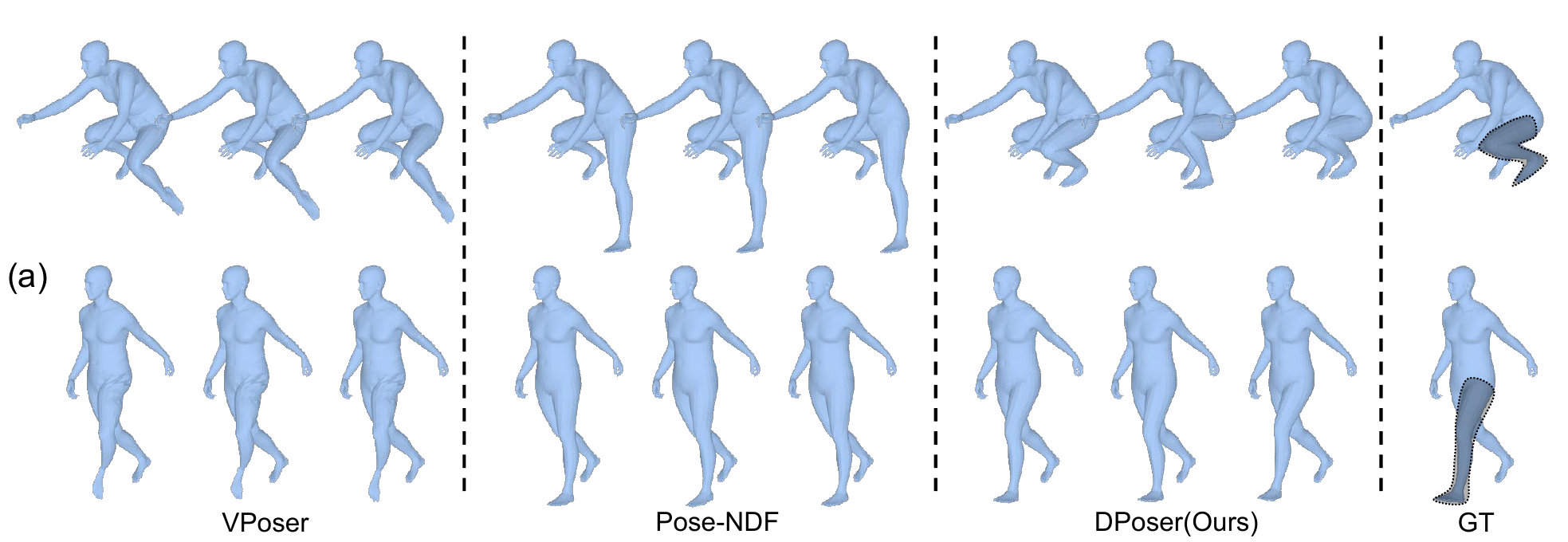}
    \end{subfigure}
    \hfill
    \begin{subfigure}[b]{\linewidth}
        \includegraphics[width=\linewidth]{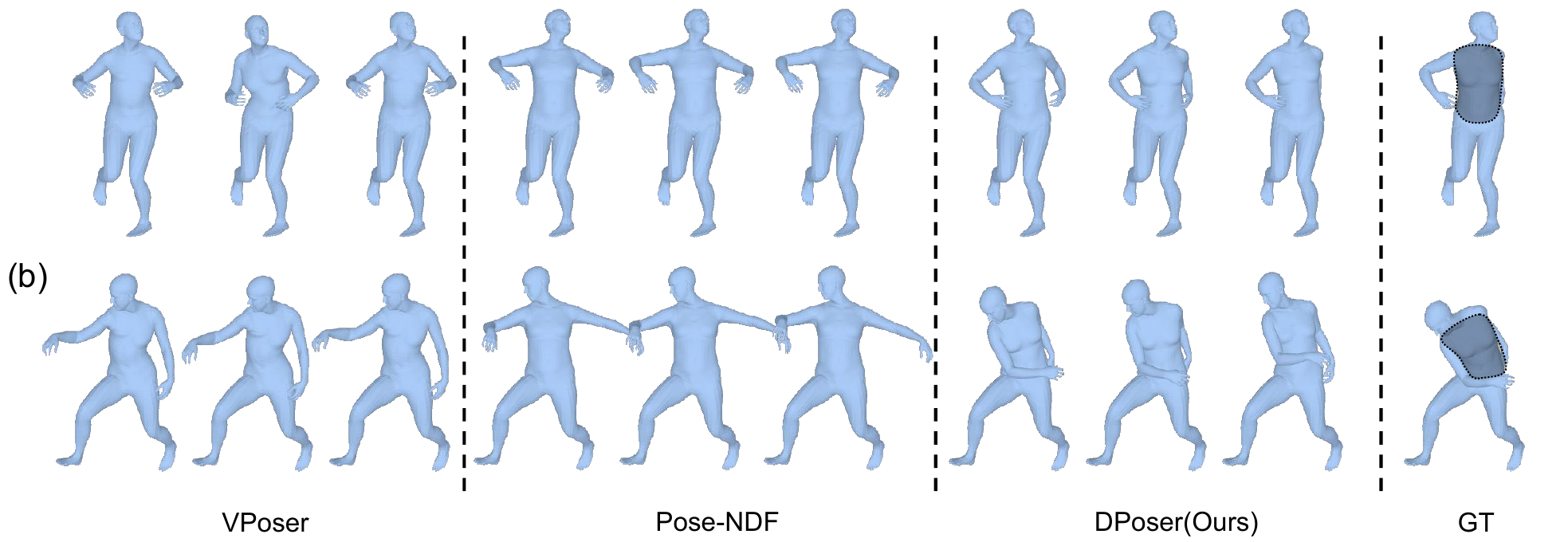}
    \end{subfigure}
    \vspace{-5.5mm}
    \caption{Pose completion. (a) Left leg under occlusion. (b) Trunk under occlusion.}
    \label{fig:more completion}
\end{figure*}

\begin{figure*}
    \centering
    \begin{subfigure}[b]{\linewidth}
        \includegraphics[width=\linewidth]{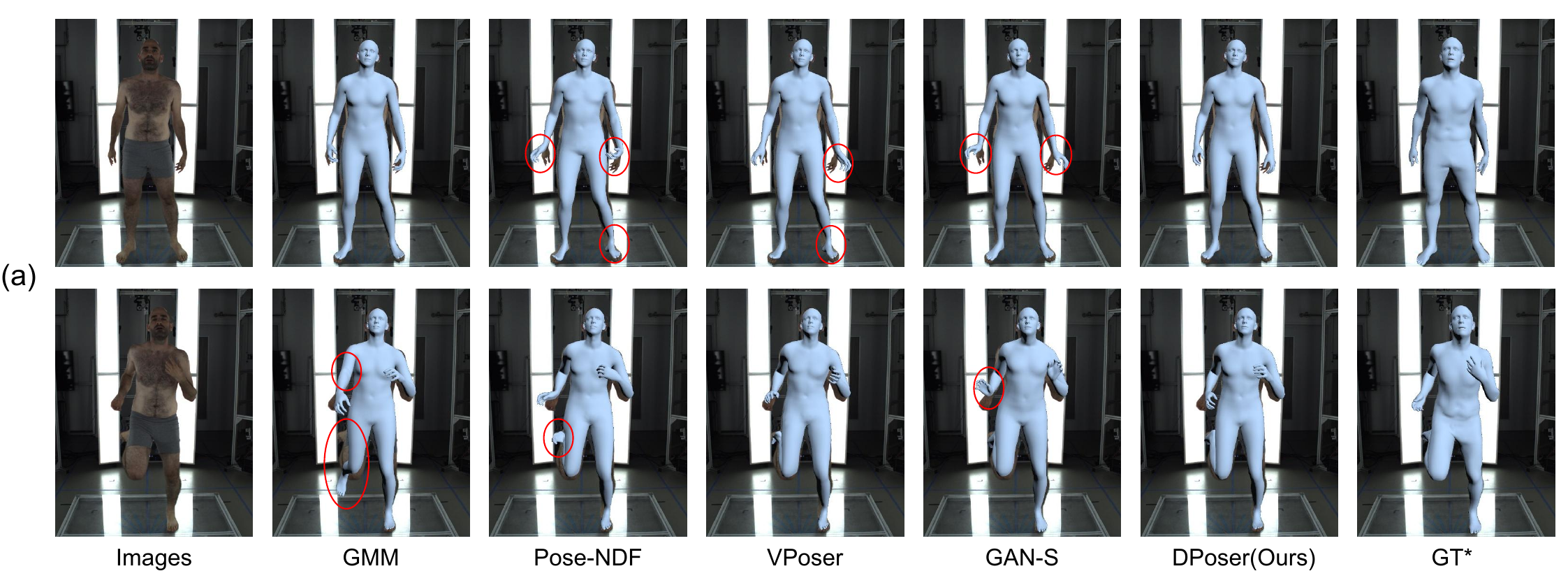}
    \end{subfigure}
    \hfill
    \begin{subfigure}[b]{\linewidth}
        \includegraphics[width=\linewidth]{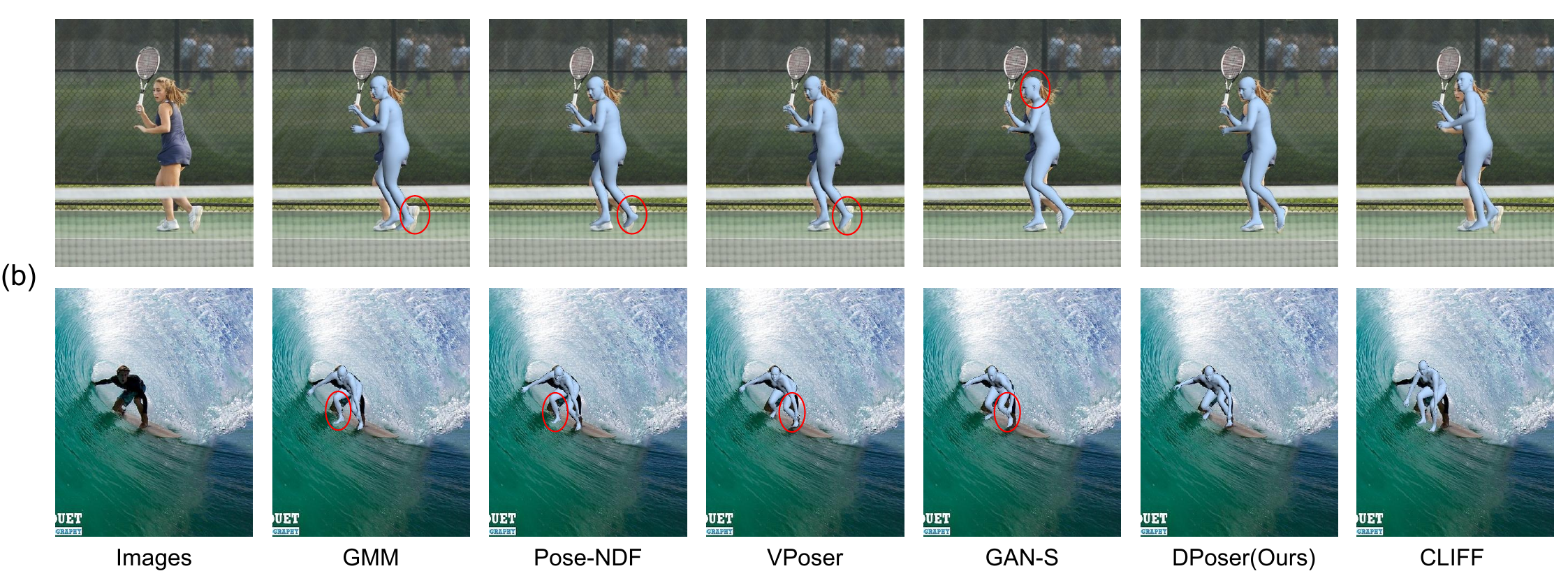}
    \end{subfigure}
    \caption{Human mesh recovery. (a) Initialization using mean poses and default camera. *Ground truth for the EHF dataset is annotated in SMPL-X~\cite{pavlakos2019expressive}, which extends SMPL~\cite{loper2015smpl} with fully articulated hands and an expressive face. (b) Initialization using the CLIFF~\cite{li2022cliff} prediction.}
    \label{fig:more hmr}
\end{figure*}

\begin{figure*}
    \centering
    \begin{subfigure}[b]{\linewidth}
        \includegraphics[width=\linewidth]{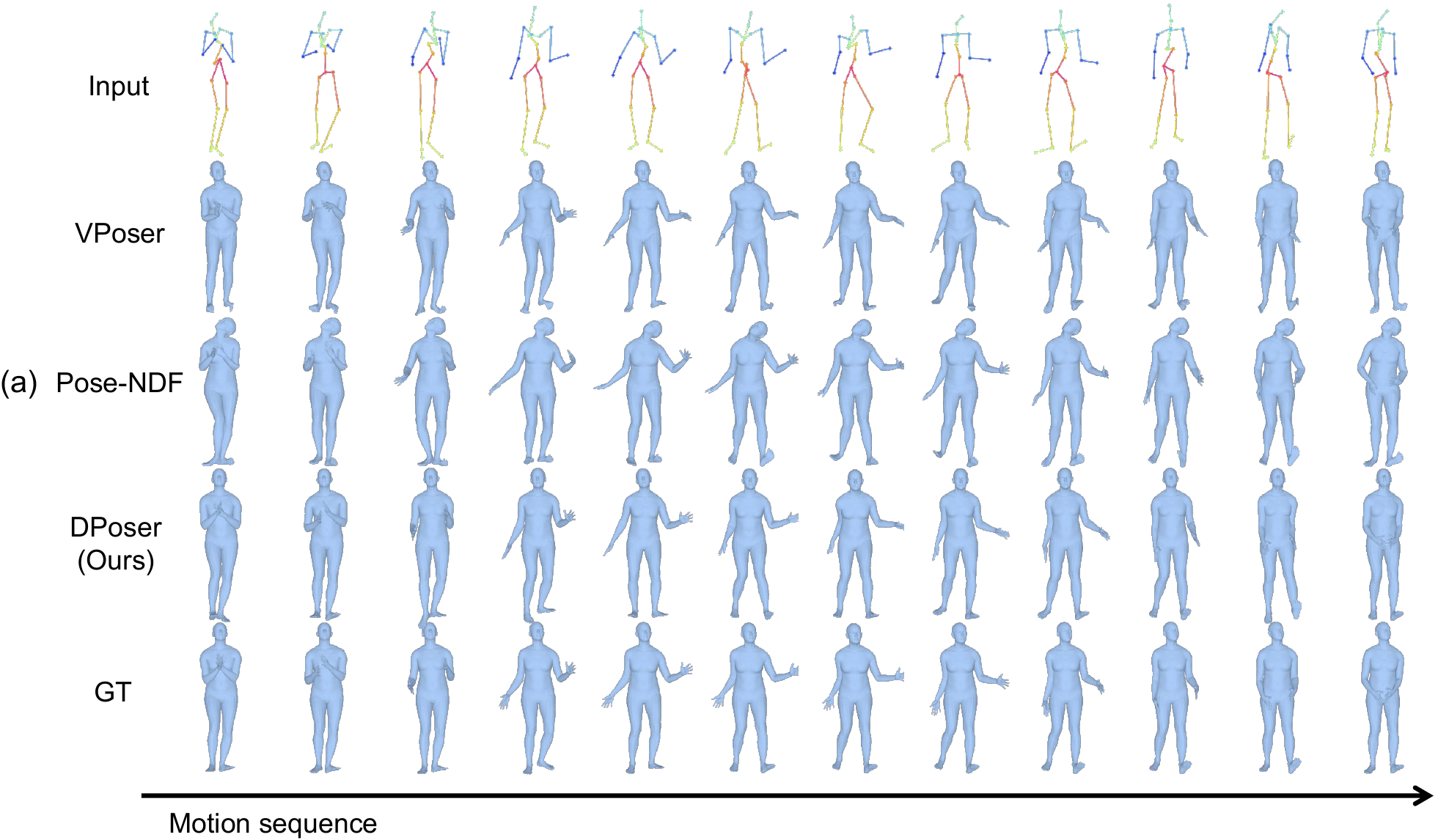}
    \end{subfigure}
    \hfill
    \begin{subfigure}[b]{\linewidth}
        \includegraphics[width=\linewidth]{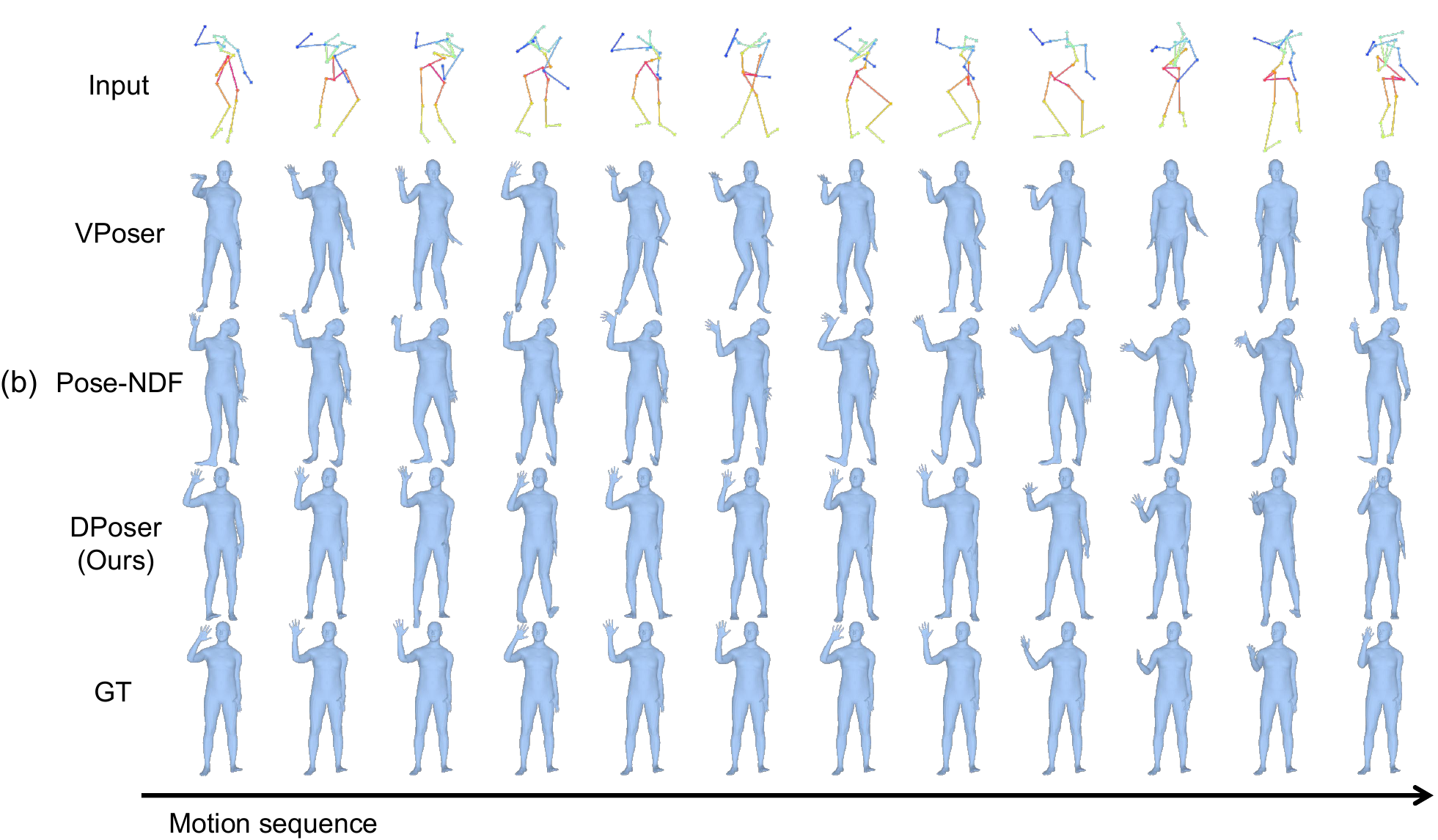}
    \end{subfigure}
    \caption{Motion denoising with noisy observations. (a) Gaussian noise with 40mm standard deviation. (b) Gaussian noise with 100mm standard deviation. We visualize every 20\(^{th}\) of the sequence.}
    \label{fig:more motion}
\end{figure*}

\begin{figure*}
    \centering
    \begin{subfigure}[b]{\linewidth}
        \includegraphics[width=\linewidth]{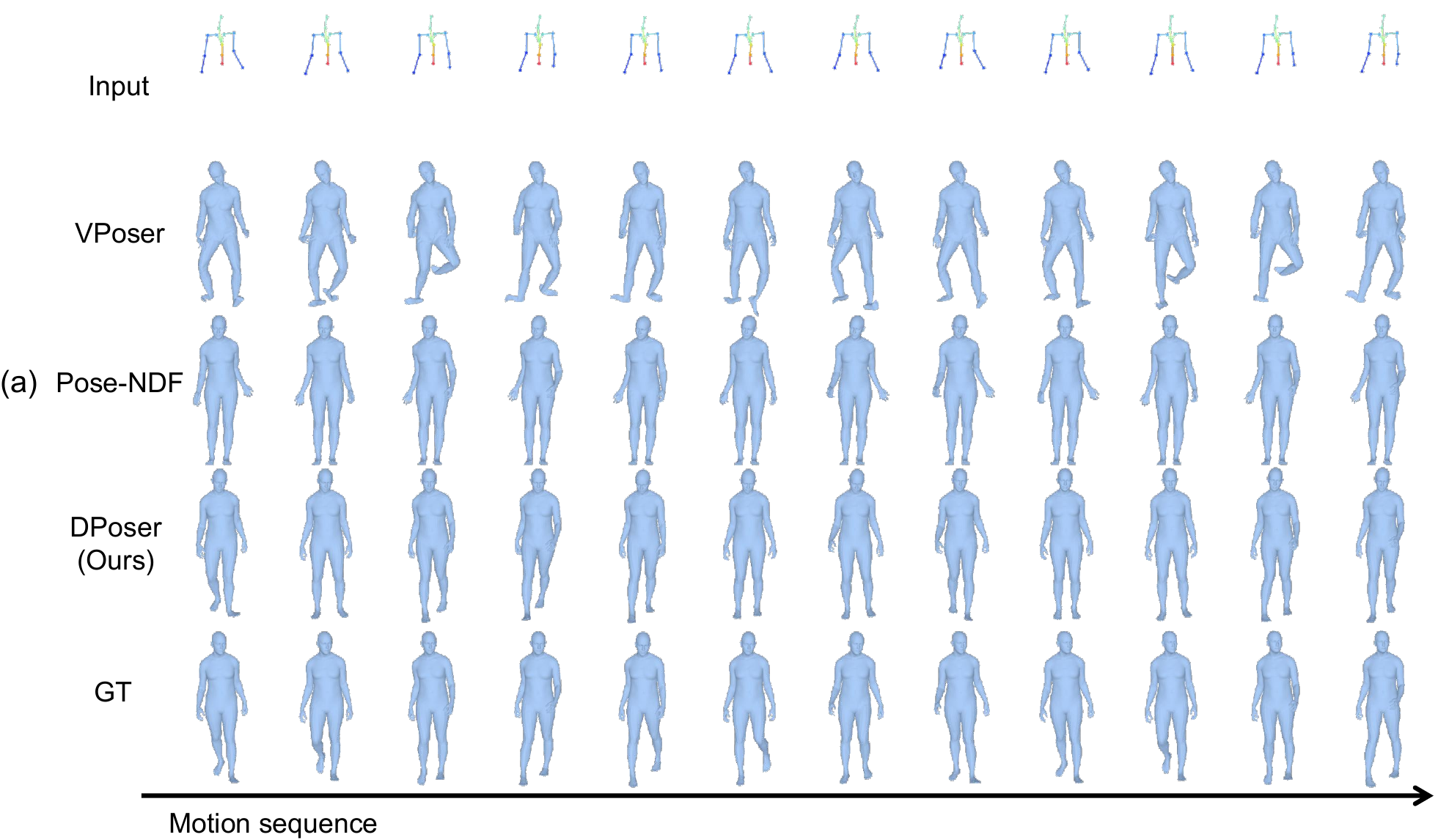}
    \end{subfigure}
    \hfill
    \begin{subfigure}[b]{\linewidth}
        \includegraphics[width=\linewidth]{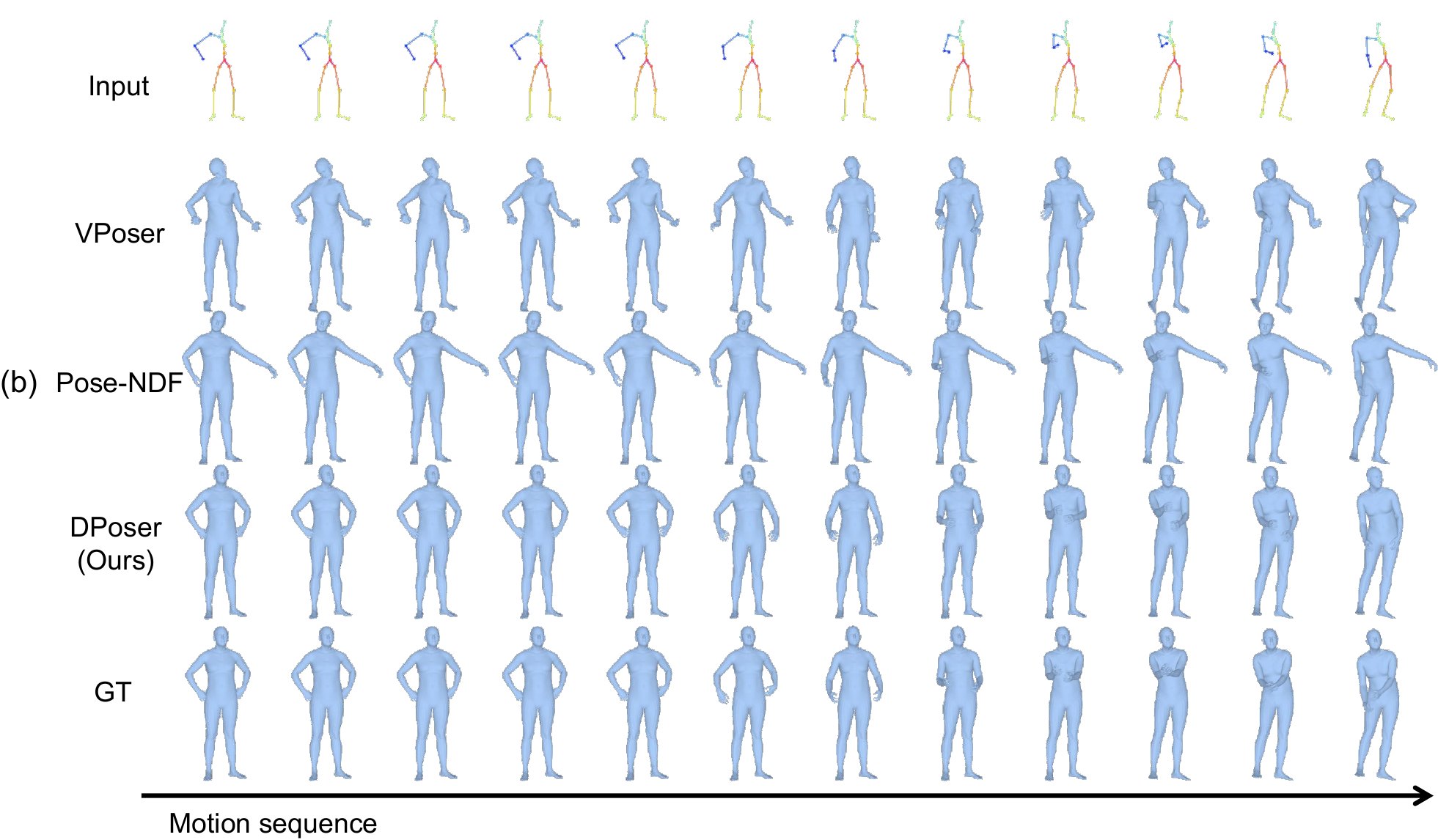}
    \end{subfigure}
    \caption{Motion denoising with partial observations. (a) Legs under occlusion. (b) Left arm under occlusion. We visualize every 20\(^{th}\) of the sequence.}
    \label{fig:motion completion}
\end{figure*}

We show more qualitative results for pose generation (\cref{fig:more generation}), pose completion (\cref{fig:more completion}), human mesh recovery (\cref{fig:more hmr}) and motion denoising (\cref{fig:more motion}, \cref{fig:motion completion}).

\section{Potential Negative Impacts}
\label{sec:ethical}
\begin{itemize}
    \item \textbf{Bias and Fairness Concerns:} Human pose prior learning models may inadvertently encode biases present in the training data, leading to biased predictions or discriminatory outcomes. This can perpetuate existing societal biases and inequalities, particularly if the training data is not representative or balanced across diverse demographics.
    
    \item \textbf{Ethical Considerations:} The use of human pose prior learning models in applications such as surveillance, security, or healthcare raises ethical concerns regarding individual privacy, autonomy, and consent. There are debates about the appropriate use of such technologies and the potential for unintended consequences or misuse.
    
    \item \textbf{Dependency on Data Quality:} Human pose prior learning models heavily rely on the quality and diversity of the training data. Poorly annotated or biased datasets can negatively impact the performance and reliability of these models, leading to inaccurate or unreliable predictions.
\end{itemize}

\end{document}